\documentclass{article} 
\usepackage{iclr2027_conference,times}

\usepackage{amsmath,amsfonts,bm}

\def\eqref#1{equation~\ref{#1}}

\def\1{\bm{1}}

\DeclareMathAlphabet{\mathsfit}{\encodingdefault}{\sfdefault}{m}{sl}
\SetMathAlphabet{\mathsfit}{bold}{\encodingdefault}{\sfdefault}{bx}{n}

\usepackage{hyperref}
\usepackage{url}

\usepackage[utf8]{inputenc} 
\usepackage[T1]{fontenc}    
\usepackage{hyperref}       
\hypersetup{colorlinks,allcolors=black}
\usepackage{url}            
\usepackage{booktabs}       
\usepackage{amsfonts}       
\usepackage{nicefrac}       
\usepackage{microtype}      
\usepackage{xcolor}         
\usepackage{amsmath}
\usepackage{graphicx}
\usepackage{subcaption}
\usepackage{tabularx}
\usepackage{wrapfig}
\usepackage{xcolor} 
\usepackage{listings}
\usepackage{multirow}
\usepackage{makecell}
\usepackage{longtable}
\usepackage{enumitem}
\usepackage{caption}

\newcommand{\code}{\url{https://github.com/MathEXLab/MMTT-Bench}}

\title{When Does Text Inform? Benchmarking Information-Theoretic Metrics for Multimodal Time-Series Forecasting}

\author{%
  Emma Andrews\\
  National University of Singapore\\
  \texttt{emma\_andrews@u.nus.edu} \\
  \And
  Gianmarco Mengaldo \\
  National University of Singapore\\
  \texttt{mpegim@nus.edu.sg} \\
}

\iclrfinalcopy 
\begin{document}

\maketitle

\begin{abstract}
Multimodal forecasting models that combine time series with text annotations promise richer prediction through textual context, but how do we know whether a text annotation meaningfully contributes to the forecasters prediction? This is an information-theoretic question, but to evaluate whether information-theoretic metrics can reliably measure the predictive value an annotation provides, a ground truth benchmark is needed, and none currently exist.
We create a synthetic time series signal with annotations in three categories: semantically \emph{correct}, \emph{incorrect}, and \emph{irrelevant}. Because the data generation process is fully controlled, ground-truth information content is known exactly, enabling principled evaluation of six complementary mutual information estimators (KSG, MINE, InfoNCE, CCA, PID and V-information). We show that all six estimators identify correct annotations as most informative, and are able to audit the quality of mixed text corpora, choosing the annotations that result in the best downstream forecasting results without the need for model training.
Our benchmark identifies limitations of each estimator, and these are validated on seven real-world datasets, which show how estimator performance differs on weak signals. Finally, we establish practical rules for implementing these metrics for annotation auditing and fusion selection.
\end{abstract}

\section{Introduction}
\label{sec:introduction}
Multimodal time series forecasting - combining numerical signals with natural language annotations - has emerged as one of the most promising frontiers in predictive modeling. 
The integration of language models with time series architectures is increasingly common across domains from healthcare~\citep{medtsllm} to climate forecasting~\citep{cllmate, kim2025comprehensive}, yet this progress rests on the key assumption that the text being fused actually helps.

In practice, this is rarely the case. 
Text may describe an upcoming regime change that the signal alone cannot anticipate, or it may be entirely wrong, or not relevant to the task at hand. 
A practitioner has no principled way to know, before training, which annotations are driving performance gains and which are adding noise. 
This is not an uncommon case, but rather a frequent situation in any real-world multimodal dataset, where text relevance to any specific future value is unknown and unverified.

Without a reliable measure of text informativeness with respect to a signal, practitioners cannot audit annotation quality before committing to expensive model training, cannot identify which domains or time periods benefit from text fusion, and cannot distinguish a model that learned from text compared to one that learned despite it. 
As multimodal forecasting scales to higher-stakes applications, the cost of fusing uninformative or misleading text grows accordingly.

Information-theoretic metrics offer a natural solution. 
Mutual Information (MI) and related measures quantify statistical dependence between text and future signal values without assuming a specific model architecture, making them ideal pre-training diagnostics: compute once on the training data to estimate if text is worth fusing before model training. 
The ability to do this reliably would provide a principled annotation auditing step that would transform how multimodal time series systems are built. 
MI estimation has already shown promise for predicting multimodal model performance~\citep{liang2023quantifying, liangmultimodal}, and as training objectives to balance modality contributions~\citep{kontras2025MCR}. 
But a fundamental obstacle remains, that is: it is not possible to evaluate these MI estimators in the text-time series setting, because real-world datasets never provide ground-truth information content. 

All existing evaluations validate these metrics against human judgment, where the true information content is commonly unknown, creating a circular dependency. 
\citet{beyondnormal2023} address this issue for a continuous unimodal numerical distribution, constructing synthetic benchmarks with known ground-truth MI, but does consider multimodal inputs such as natural language, or the extent to which modalities support or contradict each other. 
Furthermore, the notion of annotation \emph{quality} -- correct, incorrect or irrelevant --  has no analogue in existing numerical distributions. 
Our contribution to fill this gap is MMTT-Bench: a synthetic benchmark with oracle ground truth that enables the first principled evaluation of MI estimators in a text-time series multimodal setting. 
Specifically:
\begin{itemize}[leftmargin=*]
    \item We construct a synthetic dataset, MMTT-Bench (Multimodal Text-Time Series Bench): A sine wave with randomly inserted constant segments, each with three labeled annotation categories (correct, incorrect, irrelevant). We validate our findings on real-world datasets, Time-MMD~\citep{timemmd}, and FinTexTS~\citep{lee2026fintexts}. Our code and dataset is released on GitHub\footnote{\code} and HuggingFace\footnote{https://huggingface.co/datasets/WhenDoesTextInform/MMTT-Bench}. 
    \item We evaluate six complementary MI estimators, demonstrating that information metrics predict downstream forecasting performance across a variety of model architectures, validating their use as pre-training diagnostic tools.
    \item We show to what extent information metrics can be used to assess the quality of text annotations across a corpus, and our results are presented as practical rules for implementing these metrics across different signals and use cases.
\end{itemize}
%

\section{MMTT-Bench Design}
\label{sec:dataset}
\begin{wrapfigure}[17]{r}{0.5\textwidth}
     \centering
     \includegraphics[width=0.48\textwidth]{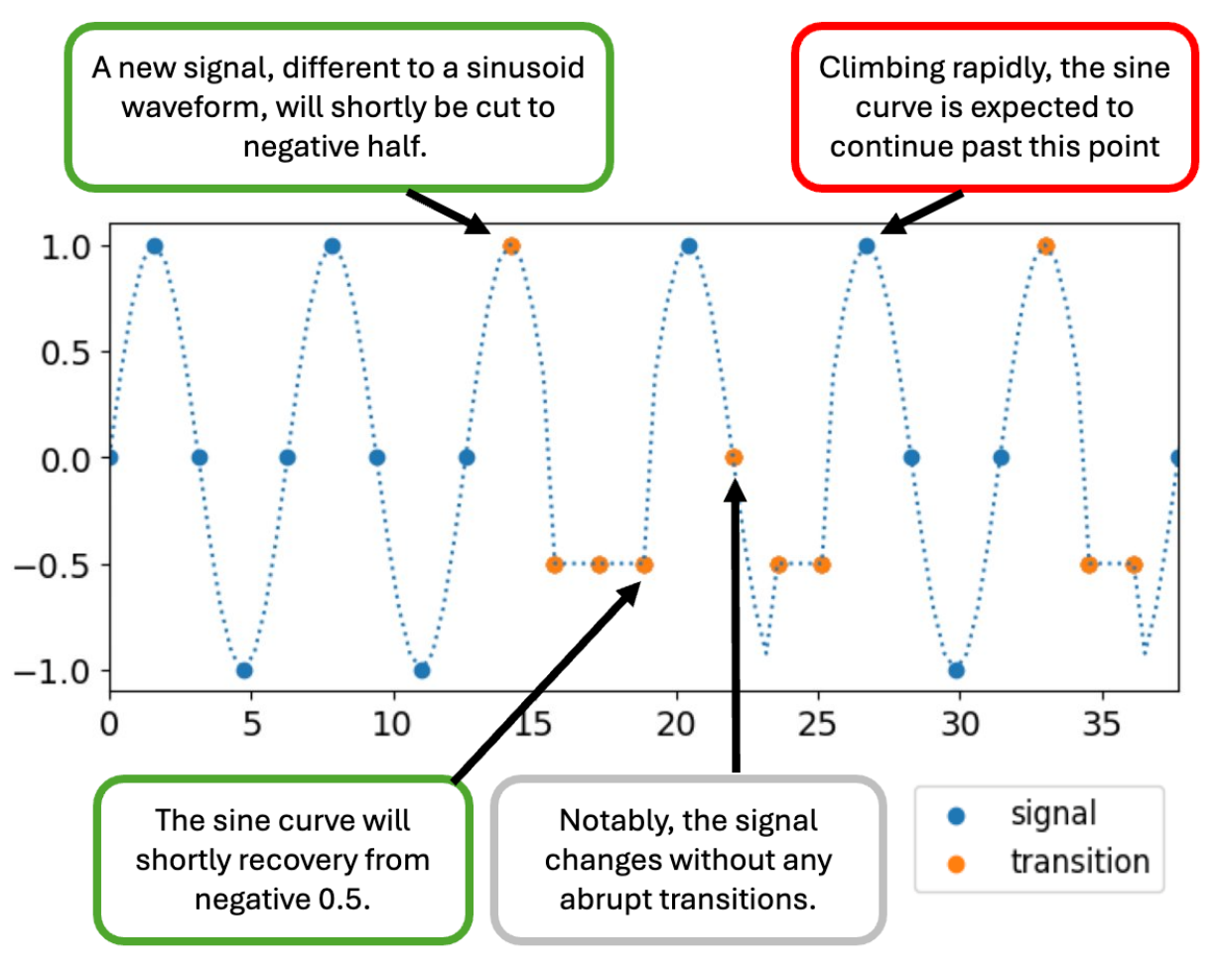}
  \caption{Example text annotations in MMTT-Bench}
  \label{fig:sinezero}
\end{wrapfigure}
We construct a synthetic benchmark, Multimodal Text-Time Series Bench (MMTT-Bench) that pairs a time series signal with controlled natural language annotations. 
Synthetic data gives us ground truth control over how much information each annotation carries about future signal values, enabling principled evaluation of information-theoretic measures across annotation categories that existing multimodal datasets do not support.

MMTT-Bench is built on a sine wave with randomly placed constant-value (flat line) segments. 
Specifically, the observable is
\[
    y(t) = \begin{cases} \sin(t) & \text{if } t \notin \mathcal{S}, \\ -0.5 & \text{if } t \in \mathcal{S}, \end{cases}
\]
where $\mathcal{S}$ is a set of randomly sampled intervals covering approximately 30\% of the time axis, each of length drawn uniformly from $[\pi/2,\ 2\pi - \pi/2]$. 
The flat line value of $-0.5$ does not coincide with any value $\sin(t)$ takes at annotations points: \(y\in[-1,0,1]\), ensuring the two regimes are unambiguous. This creates ``transition states'', depicted in orange in Figure~\ref{fig:sinezero}, where the signal drops to \(-0.5\). At these points, the time series signal values \(y(t)\) give no deterministic information about the next value of \(Y_\mathrm{future}\), and correct text annotations become the dominant source of predictive information.

The dataset is split chronologically into non-overlapping temporal segments (training \(62.5\%\) , validation \(18.75\%\),testing \(18.75\%\)) to prevent data leakage and reflect realistic forecasting settings where models are trained on historical data and evaluated on future observations. 
Each time point retrieves three annotation variants; \emph{correct}, \emph{incorrect} and \emph{irrelevant} text, based on the known future state of the underlying signal. Full dataset generation details are found in Appendix~\ref{sec:appendix-dataset-generation}.

To ensure the diversity of text is the same for the incorrect text corpus, we randomly choose these transition templates for 30\% of incorrect annotation points, equaling the number of true transition points in the signal, so that the correct and incorrect text corpus are indistinguishable without the additional time series information. 
This is confirmed in Appendix~\ref{sec:appendix-embed}, and prevents metrics from exploiting lexical cues rather than the semantic content and its relation to future predictions. 
Irrelevant text is generated from separate templates, with text describing generic properties of the signal without any directional or predictive information.
A well-calibrated information metric should rank \emph{correct} highest, \emph{irrelevant} near zero and \emph{incorrect} somewhere in-between, since 70\% of incorrect annotations provide directly opposing information to correct annotations, they are still correlated with \(Y_{\mathrm{future}}\) values. 
A model could learn to do exactly the opposite of what the incorrect annotations suggest, and be correct up to 70\% of the time.

The core question is whether MI and related metrics can distinguish and correctly order these text variants, and if the associated metrics predict the performance gain when each annotation type is given to a forecasting model.


\section{Estimating Mutual Information}
\label{sec:mutual-information}

\subsection{Notation and Background}
\label{sec:notation}
We set up the task of multimodal forecasting as follows:
At each annotation point $t$, we observe three quantities. \(X_{\text{ts}}\) is
the time-series input: a vector of the \(n_{\text{lookback}}\) preceding signal values (optionally tokenized using PatchTST~\citep{PatchTST}). \(X_{\text{text}}\) is the text input: the natural-language annotation at \(t\),
embedded and reduced by PCA to \(d_{\text{text}}\) dimensions. \(Y\) is the target: the signal value
at \(t + \text{horizon}\). Every quantity we report is a mutual information (or mutual-information-like)
functional of the joint distribution over \((X_{\text{ts}}, X_{\text{text}}, Y)\), estimated from a finite sample of annotation points.

For each annotation category, we compute four quantities: 
the marginal MI of each modality individually, \(I(X_{\mathrm{ts}};Y)\) and \(I(X_{\mathrm{text}};Y)\), which measure how much information each modality shares with \(Y_{\mathrm{future}}\), the joint MI, \(I(X_{\mathrm{ts}},X_{\mathrm{text}};Y)\), which measures the combined information of both modalities, and the conditional MI, \(I(X_{\mathrm{text}};Y | X_{\mathrm{ts}}) = I(X_{\mathrm{ts}},X_{\mathrm{text}};Y) - I(X_{\mathrm{ts}};Y)\) that isolates the unique contribution of text beyond the time series.
 This is our primary quantity of interest, because it is the quantity a practitioner actually needs before deciding whether fusing text is worth the cost.

\textbf{But why must MI be \emph{estimated?}} For continuous, high-dimensional variables with unknown
densities, mutual information has no closed form: computing it exactly would require integrating
over the true joint density $p(X_{\text{ts}}, X_{\text{text}}, Y)$, which is never available in
practice. This is true even in our synthetic setting, despite the information content of an
annotation being known by construction, the mapping from raw text through an embedding model and PCA to a usable representation is not analytically tractable. Every number we report is therefore an estimate with its own bias and variance properties, not a ground
truth value. This is precisely why an oracle benchmark that separates estimator error from
unknown-true-value uncertainty is necessary. MI estimation is specifically difficult in the text–time-series setting, motivating the need to quantify the performance and limitations of different estimators.


\subsection{Estimators}
We evaluate six complementary MI estimators spanning the full space of approaches to measuring text-time series information content. 
K-Nearest-Neighbor Estimator (KSG)~\citep{ksg2004}, Mutual Information Neural Estimator (MINE)~\citep{mine2018}, InfoNCE~\citep{infonce2024} and Canonical Correlation Analysis Estimator (CCA)~\citep{murphy2023} all target the same Shannon MI quantity \(I(X;Y)\) in nats, and are directly comparable in scale but differ in assumptions and practical behavior.
Partial Information Decomposition (PID)~\citep{williams2010} decomposes the joint information directly into non-negative redundancy, unique, and synergy atoms, so the conditional quantity of interest is its
\(\text{Unique}(X_{\text{text}})\) atom rather than a subtraction. V-Usable Information (V-Information)~\citep{xu2020} departs from Shannon entropy entirely, measuring the information a Ridge regression can exploit from the given embedding, reported as \(\Delta R^2\), the gain in explained variance from adding \(X_{\text{text}}\) to a fixed predictive family. Full derivations and implementation details are provided in Appendix~\ref{sec:appendix-metrics}.

These estimators all have limitations which we aim to quantify in the text-time series settings.
Estimator variance and the required sample size both scale with the magnitude of the quantity being estimated, so weak-signal regimes where text contributes limited additional information are intrinsically harder to estimate. This is especially problematic for KSG, which estimates density locally via nearest-neighbor distances, and therefore becomes unreliable once the joint dimensionality \(d\) exceeds exceeds \(\sqrt{N/2}\) for sample size \(N\) \citep{gao2017}. This motivates dimensionality reduction of \(X_{\text{text}}\), (see Section~\ref{sec:representation}).  Variance for KSG is estimated using the recommended procedure of~\citet{holmes2019estimation} rather than naive bootstrapping, which is used for all other estimators. 

MINE and InfoNCE learn a critic network by gradient descent and are insensitive to dimensionality, but require hyperparameter tuning and sufficient signal strength to train reliably. 
They also produce bounds rather than true estimators. MINE is a lower bound obtained via the Donsker-Varadhan representation~\citep{mine2018}, and InfoNCE is a lower bound that saturates at \(logN\) nats for batch size \(N\)~\citep{infonce2024}. Both can therefore systematically underestimate true MI.

 Estimating a conditional as a difference of two separately biased estimates can leave a residual larger than the quantity being measured. Because
each term carries its own estimator-specific bias, the two biases do not cancel in general. When the marginal
\(\hat{I}(X_{\text{ts}}; Y)\) is large relative to the true conditional MI, this bias can dominate the difference and produce a negative estimate for a quantity that is non-negative by definition. This can cause negative KSG conditional MI values, and is one advantage of CCA, whose chain rule holds exactly under its Gaussian assumption, so its conditional estimates do not exhibit this inconsistency~\citep{beyondnormal2023}.
\subsection{Related Work}
\label{sec:related-work}

The use of mutual information to characterize multimodal data has gained significant traction across two complementary directions; as an analytical tool to predict model behavior before training, and as a training objective to improve model performance.

On the analytical side, \citet{liang2023quantifying} introduce a PID framework and show it can not only characterize dataset structure but also predict the performance of downstream multimodal models and guide model selection, all without training a single model. 
A concurrent line of work by~\citet{liangmultimodal} extends this in a semi-supervised setting. 
However, both approaches validate estimation quality against human judgment on real-world datasets where the true data generation process is unknown, creating a circular dependency between what PID captures and what humans intuitively expect it to mean. 
Our benchmark addresses this directly by providing oracle ground truth through control of the data generation process, so the information-theoretic quantities can be validated without recourse to human annotation.

On the training side, \citet{kontras2025MCR} introduce the Multimodal Competition Regularizer (MCR), which uses a single conditional MI estimate as a training signal to adaptively balance modality contributions, encouraging each modality to maximize its unique predictive role. MCR optimizes this quantity during training rather than evaluating it, and uses a single estimate per training step rather than comparing estimators. Since the quantities MCR relies on - unique and shared modality contributions - are identical to those our benchmark measures, our results directly inform which estimator MCR should use and under what signal conditions its loss can be trusted.

\citet{wang2026ragen}, similarly uses MI as an online training diagnostic during reinforcement learning of LLM agents, finding that mutual information between input prompts and generated reasoning traces correlates with final task performance much more strongly than entropy alone. 
These applications demonstrate the practical demand for reliable MI estimation, yet both validate their MI-based signals based on model performance alone, when true information content is unknown. Without a benchmark that provides an oracle ground truth for MI estimation, there is no principled way to measure if model performance improvements are driven by accurate MI estimation or occur in spite of it.

A complementary perspective is offered by \citet{lee2026rethinking} and \citet{zhang2025does}, who show empirically that naive multimodal fusion of text and time series frequently under performs unimodal baselines due to uncontrolled integration of irrelevant textual information. This motivates the need for principled measurement of textual relevance before fusion.
This also demonstrates a problem with existing multimodal time series benchmarks: they do not provide the ground truth labels or controlled annotation quality necessary to evaluate information-theoretic metrics.


While Time-MMD~\citep{timemmd} initially demonstrated adding text achieves MSE reduction, more recent work achieves better forecasting results using only the time series, making us question if the text is truly informative~\citep{zhang2025does}. FinTexTS~\citep{lee2026fintexts} claims to improve the text quality of FNSPID~\citep{dong2024fnspid}, a stock price dataset that is paired with news articles, by replacing the keyword-based matching strategy with a semantic and multi-level pairing framework. It claims the new pairings result in better forecasting performance than no text and original paired text baselines~\citep{lee2026fintexts}. 
Context is Key~\citep{cik} provides human-authored context descriptions where the text is strictly necessary to predict the future signal, however it contains a relatively limited number of annotated points, 400 across 14 tasks, making it too small for reliable MI estimation. 
None of these benchmarks provide the structure of  correct, incorrect and irrelevant annotations that are necessary to establish ground truth for information-theoretic measures of text quality.

The closest precedent is~\citet{beyondnormal2023}, who construct a diverse family of synthetic distributions with analytically known ground-truth MI and uses them to systematically benchmark estimator accuracy. 
We adopt the same principle, with the key distinction is that~\citet{beyondnormal2023} operate on pairs of continuous random numerical variables with no semantic structure. 
Our benchmark, instead, introduces natural language as one modality. 
The extension is non-trivial, text must be embedded before MI estimation, introducing a representational bottleneck that does not exist in purely numeric settings, and the notion of annotation quality -- correct, incorrect or irrelevant -- has no analogue in continuous synthetic distributions. 

\subsection{Representation Pipeline}
\label{sec:representation}
\begin{figure}[h]
  \begin{subfigure}[t]{0.5\textwidth}
         \centering
          \includegraphics[width=\textwidth]{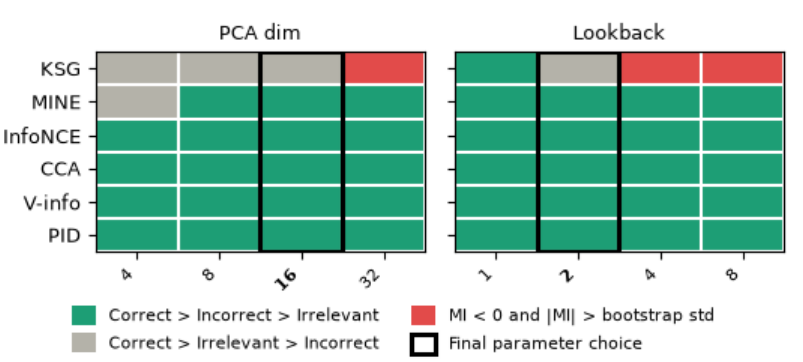}
     \end{subfigure}
     ~ 
     \begin{subfigure}[t]{0.4\textwidth}
         \centering
          \includegraphics[width=\textwidth]{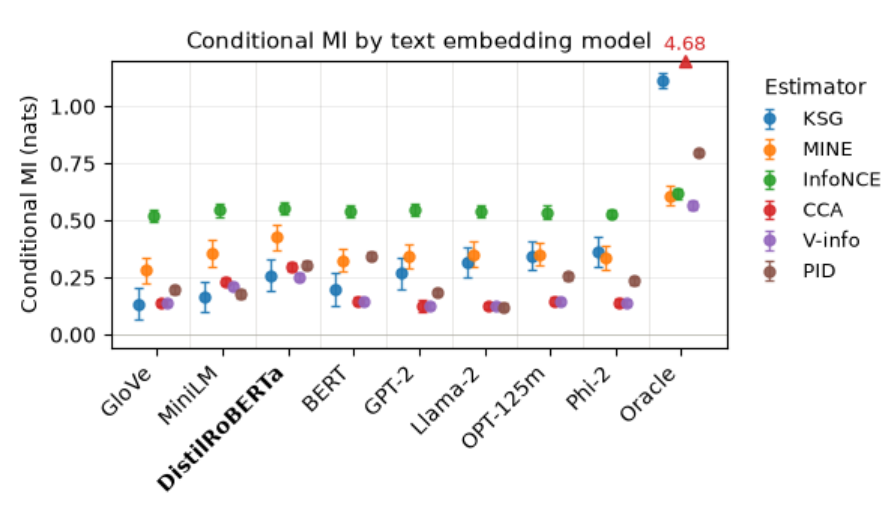}
     \end{subfigure}
  \caption{Results of parameter sweeps on MMTT-Bench. Left plot shows how PCA dimension and lookback steps affect MI ordering. Right plot shows conditional MI estimates for correct text annotations using different text embedding models.}
  \label{fig:representation}
\end{figure}

To ensure the estimator comparisons are robust to different representation pipeline choices, Figure~\ref{fig:representation} compares how the PCA dimension and number of lookback steps affect the ordering of MI estimates across text categories, where we expect correct > incorrect > irrelevant. As we expect MI to be close to zero, we define an estimate as negative if the mean across seeds if less than zero and the magnitude is greater than the bootstrap standard deviation of that estimator.
KSG collapses to produce negative values as it approaches \(\sqrt{N/2} \approx25\) bound as predicted, hence we choose \(d_{\text{text}}=16\), \(n_{\text{lookback}}\leq 2\) as our final parameters. Lowering PCA dimension further causes KSG and MINE to estimate MI for irrelevant text above incorrect, which we know not to be true, therefore we assume reducing the text this dramatically loses information that helps differentiate between these two text annotations.

Figure~\ref{fig:representation} also shows how conditional MI remains relatively stable across all estimators when varying the text encoder. This holds across eight embedding models spanning sentence transformers and LLM encoders. We also include an 'oracle' embedding which simply includes the next \(Y_{\text{future}}\) value, representing an upper bound if the correct text information was embedded perfectly. The ordering of MI estimates across different text categories is preserved in all cases.
DistilRoBERTa is selected because it balances high \(I(X_\text{text};Y)\) relative to \(I(X_\text{ts};Y)\) for low computational cost, indicating it best preserves semantic content relevant to \(Y_\mathrm{future}\) within the PCA-reduced representation. 
\begin{figure}
 \centering
  \includegraphics[width=0.9\textwidth]{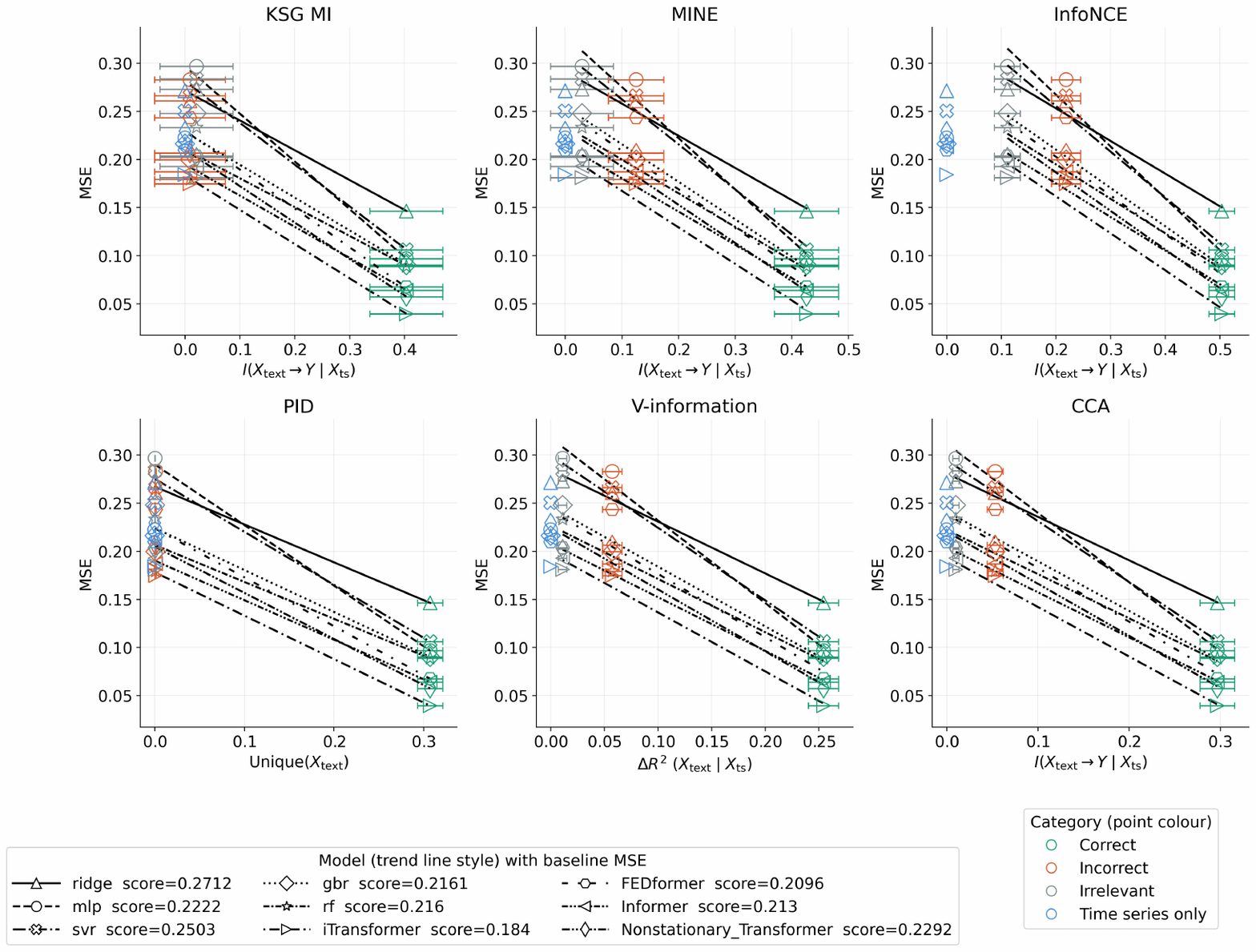}
  \caption{Information Metrics measuring text contribution compared to change in model performance between \(X_{joint}\) and \(X_{ts}\) for MMTT-Bench}
  \label{fig:mmtt_bench_sklearn}
\end{figure}
\section{Results}
\label{sec:results}
In our results, corpus text quality is in the independent variable, constructed as three categories, correct, incorrect, and irrelevant. The MI metric is the first dependent value, does the estimator's estimate change with corpus quality? This is our estimator evaluation. Finally downstream model forecasting performance, measured through mean squared error (MSE), is the second dependent variable. This is only presented to establish that the ground-truth quality ordering is consequential; a corpus labeled as better really does train a better model. We train fourteen model architectures and, where appropriate, ten different fusion mechanisms per architecture, so this result does not depend on a single model's inductive biases.

Figures~\ref{fig:mmtt_bench_sklearn} displays model forecasting performance (MSE) on the y-axis and conditional MI estimates of \(X_\text{text}\) for each information metric on the x-axis.
Conditional MI of \(X_\text{text}\) represents the unique information the text annotations contribute to \(Y_\mathrm{future}\) predictions. For KSG, MINE, InfoNCE and CCA this is \(I(X_\text{text} \rightarrow Y | X_\text{ts})\), for PID it is Unique(\(X_\text{text}\)), and for V-Information it is the change in ridge regression performance \(\Delta R^2\) when trained with \(X_\text{text}\) and \(X_\text{ts}\) minus the performance when trained only on \(X_\text{ts}\).

\subsection{MMTT-Bench}
\label{sec:results-sine-wave}
Metrics were computed on the training split consisting of \(N=1279\) annotation points, each with correct, incorrect and irrelevant text. Full parameter choices and sweep results are reported in Appendix~\ref{sec:appendix-sine-sweeps}, and results for every metric quantity, not just conditional MI, are in Appendix~\ref{sec:appendix-sine-final}.

The y-axis in Figure~\ref{fig:mmtt_bench_sklearn} shows the model performance across nine model architectures. For simplicity we only included the best performing models, with the four time series transformers using a basic additive fusion method. Appendix~\ref{sec:appendix-sine-final} provides full results, including a further four transformer architectures and 10 different fusion strategies.
Firstly, focusing on the MI estimates across different text categories, Figure~\ref{fig:mmtt_bench_sklearn} confirms that, on a well-designed benchmark with sufficient data and appropriate dimensionality, correct text is reliably identified as the most informative category.
However, KSG and PID assign near-identical values to incorrect and irrelevant, despite it being possible to separate them by design.
All other estimators rank incorrect \(>\) irrelevant with statistical significance, and we note that larger error bars are expected when bootstrapping neural estimators MINE and InfoNCE because it reflects both sampling uncertainty and training stochasticity.

This rank reflects a genuine property of the data. Incorrect annotations describe the \emph{opposite} future direction to the true signal for 70\% of points, 
and therefore carries real statistical dependence with \(Y_{\mathrm{future}}\) under an inverted mapping, while irrelevant text encodes no dimension systematically related to \(Y_{\mathrm{future}}\) in any direction. 
This is validated in embedding space, where irrelevant annotations form a distinct cluster, and correct and incorrect overlap almost completely (see Figure~\ref{fig:sine_embed}, Appendix~\ref{sec:appendix-embed}).
Neural estimators detect this to a greater extent because their critics learn \(p(Y\mid X_\text{text})\) directly through contrastive scoring, capturing any consistent mapping regardless of sign; non-neural estimators summarize averaged co-variation, making it harder to resolve the distinction.

In Appendix~\ref{sec:appendix-shuffle} we confirm that the incorrect signal is carried by the text-time series pairing rather than the text alone by shuffling incorrect annotations across time points and recomputing all metrics. MINE and InfoNCE both shows a significant drop in MI estimates of \(\approx80\%\) for correct text and \(\approx30\%\) for incorrect text after shuffling, with no change in irrelevant text, confirming that the inverted signal is a consistent property of the (text, \(Y_{\mathrm{future}})\) pairing and not an artifact of embedding distribution.
\begin{wrapfigure}[25]{r}{0.5\textwidth}
     \centering
      \includegraphics[width=0.48\textwidth]{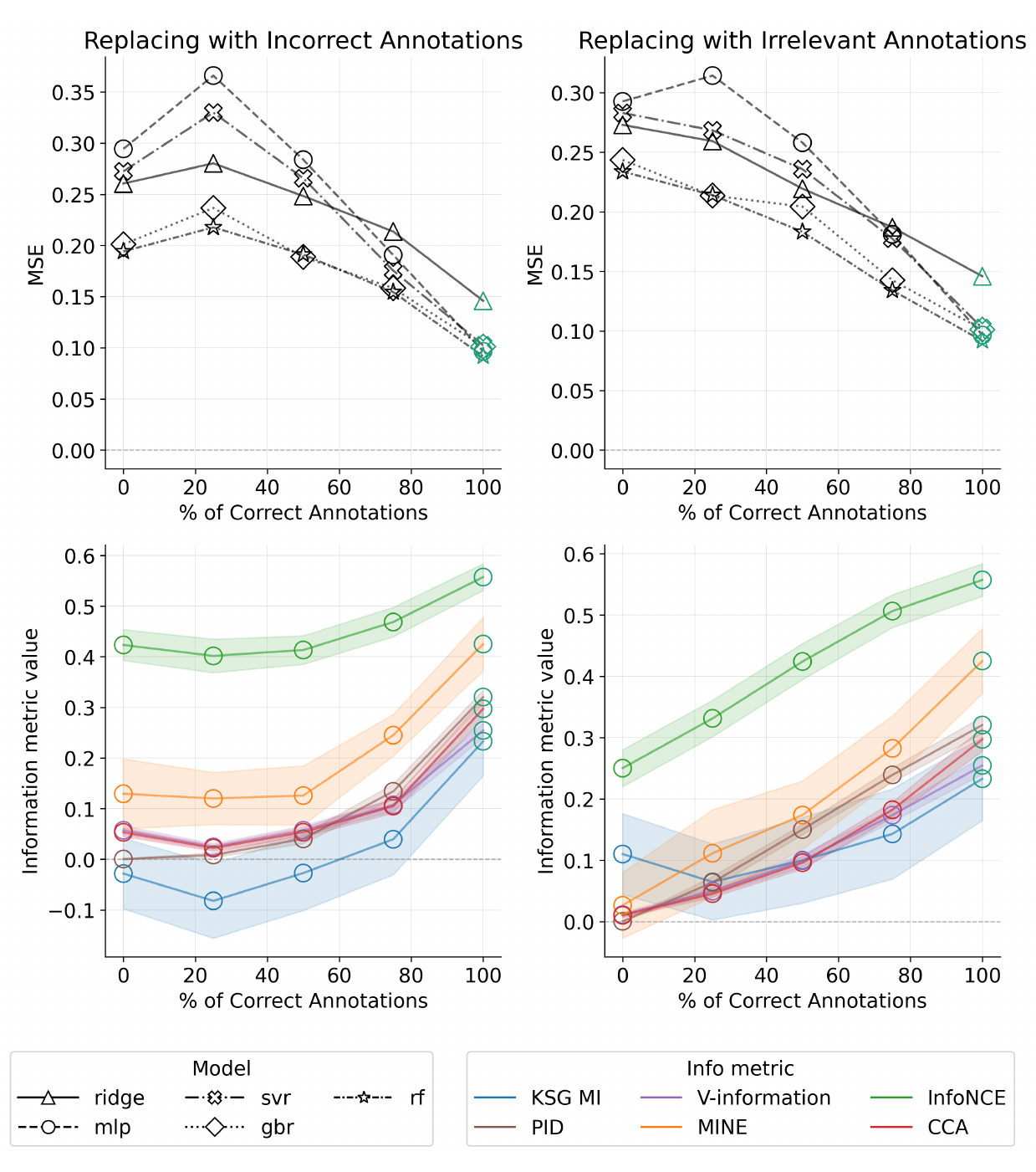}
  \caption{Model Performance (top) and Information Metrics (bottom) compared to Percentage of Correct Annotations}
  \label{fig:mixture}
\end{wrapfigure}
%
Focusing on the y-axis (model performance), Figure~\ref{fig:mmtt_bench_sklearn} shows a consistent positive relationship between metric value and downstream performance across all six estimators, with trend lines well-separated by model baseline but aligned in slope. The time series only baseline (blue markers) anchors the left of each panel at zero MI from the addition of \(X_\text{text}\) by default, clearly showing the metric values correctly attribute performance gains to text rather than time series.

Figure~\ref{fig:mixture} extends this to the full quality spectrum by varying the fraction of correct annotations from 0\% to 100\%, replacing the remainder with either incorrect or irrelevant text.
Under irrelevant replacement (right panels), both metric values and \(MSE\) change monotonically across all estimators and architectures, confirming that metrics are well-calibrated continuous sensors of corpus quality. 
Under incorrect replacement (left panels), model performance shows a U-shape: at 0\% correct, entirely incorrect annotations still sustain partial model performance via the inverted signal, before degrading as the mix becomes ambiguous and recovering as correct annotations dominate. Metric values mirror this pattern, demonstrating that information metrics track the true predictive value of a corpus even in cases, such as the inverted signal, that human annotation quality 
We use the mixture dataset's nine corpora of varying qualities to demonstrate that conditional MI estimates can be used as a pre-training audit score to select the corpus which, when used for training, results in the best performance. Across the mixture corpora and five simple downstream models, Spearman \(|\rho|\) between conditional MI and MSE is 0.849 on average, 28/36 estimator-model pairs exceed \(|\rho| > 0.8\) and 32/36 are significant with \(p<0.05\). CCA is the highest and best auditor (0.90-1.00) and KSG lowest (0.40-0.73). 
Every estimator's top MI is the 100\% correct corpus, which is also the best model performance across all models, giving a performance boost of \(-0.114\) MSE compared to selecting text annotations at random from correct, incorrect and irrelevant, and beating the no text baseline. However as selecting a clean corpus is not a hard test, we remove the corpora with all annotations from the same category, and every estimator selects the corpus contaminated with 25\% irrelevant text, which is also the best model performance for 5 out of 6 models, and gains \(-0.07\) MSE over a random selection. 
In Appendix~\ref{appendix:signal-perturbations} we further demonstrate these results hold when adding noise to the sine signal and introducing controlled perturbations in the form of temporal jitters, where each annotation is replaced by the correct text of a nearby time point.

\subsection{Real-World Validation}
\label{sec:time-mmd}
To investigate whether the information-theoretic findings from our synthetic benchmark transfer to real-world data, we extend our evaluation to seven domains from Time-MMD~\citep{timemmd} and ten stocks from FinTexTS~\citep{lee2026fintexts}. Appendix~\ref{sec:appendix-real-world-datasets} shows that these span trend, seasonality, non-stationary and multivariate structure, everything the sine signal lacks. 
\begin{wrapfigure}[18]{r}{0.6\textwidth}
     \centering
      \includegraphics[width=0.58\textwidth]{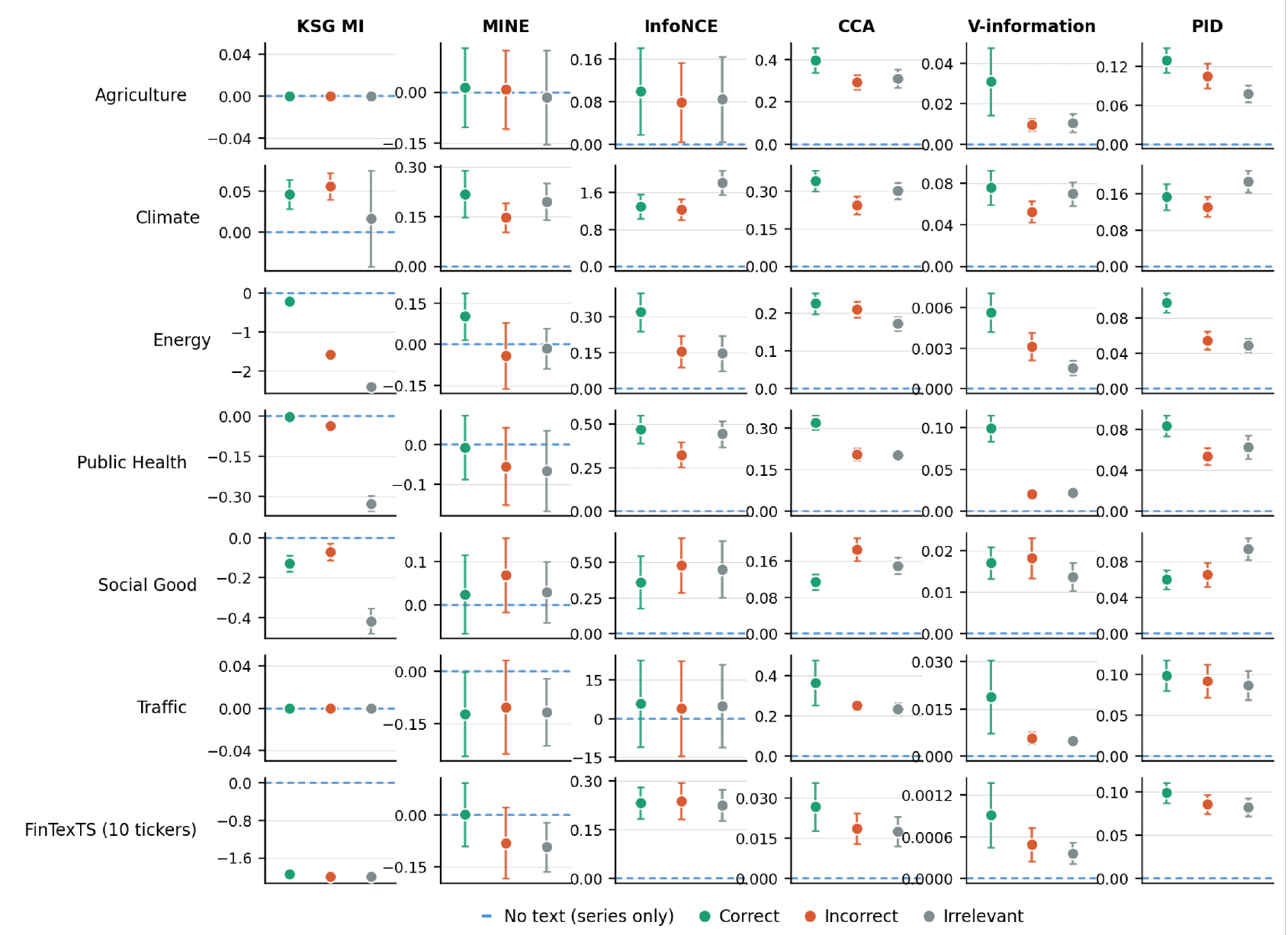}
  \caption{MI Estimates on seven real-world datasets}
  \label{fig:real_grid}
\end{wrapfigure}
Baseline time-series predictability is substantially higher here for Social Good, Public Health and Energy on time series alone (\(R^2\) in [0.65-0.93[)] compared to 0.37 on MMTT-Bench. 
This limits the room available for text to contribute and makes this a meaningfully different, and harder, regime than the synthetic setting. Despite this, every estimator pathology characterized under controlled conditions on MMTT-Bench reappears here: KSG's conditional MI turns negative where the marginal-to-conditional ratio is large, MINE turns negative and InfoNCE inflates under weak signal, and CCA, V-information and PID remain stable and correctly ordered throughout. 

Similar to the audit test carried out on MMTT-Bench, we also seek to evaluate estimators for practical applications. One example is rather than simply identifying whether text is informative in aggregate, a practitioner also needs to know whether a given
annotation is trustworthy \emph{for the timestamp it is attached to}. We distinguish these with two bootstrapped contrasts (20 resamples; \(z = \Delta/\sigma\)): the conditional MI to measure whether text is informative as before, and an alignment contrast between correctly-paired text and same-domain text drawn from the wrong timestamp, which isolates whether the \emph{pairing} carries information. Across the seven datasets, conditional MI is positive and significant while the alignment contrast is statistically indistinguishable from zero for most estimator–dataset pairs (Appendix~\ref{sec:appendix-real-results}, Table~\ref{tab:real_align}): text in these corpora carries real information, but that information is largely diffuse and topical rather than tied to the specific date it is attached to - despite every one of these datasets being constructed on the assumption that it is date-specific. Downstream model performance in Appendix~\ref{sec:appendix-real-results}, Table~\ref{tab:real_performance} corroborates this: when adding text, median MSE gets worse for every dataset except Climate, and on Agriculture and Public Health, irrelevant text degrades performance \emph{less} than correct text. This is the empirical basis for treating conditional MI and the alignment contrast as answering two different questions.
 \section{Recommendations}
 \label{sec:recommendations}
 \begin{enumerate}[leftmargin=*]
 \item \textbf{Default audit pair: CCA and V-information.} They are deterministic given a resample, require no training, are scale-invariant, and satisfy the chain rule (exactly for CCA under the Gaussian assumption), never producing a negative conditional MI. On the mixture corpora their conditional MI rank-correlates with downstream MSE at \(|\rho| =\)0.90-1.00 across five model families, the highest of any estimator.

\item \textbf{Use neural estimators (MINE, InfoNCE) only when N is large and the expected conditional MI is substantial.} These estimators reliably separated incorrect from irrelevant text, because their critics learn \(p(Y|X_{\text{text}})\) directly and are therefore sensitive to sign-inverted dependence, which averaged co-variation summaries cannot see. But on real world data, where conditional MI is small, MINE collapses to near-zero and InfoNCE shows very high variance, even after hyperparameter tuning. This shows they are unreliable when the signal provided by the text is weak, or N is small.

\item \textbf{Do not use KSG for conditional MI when the marginal \(I(X_{\text{ts}};Y)\) is large.} Because the conditional is a difference of two separately-biased estimates and KSG's bias depends on dimensionality, a large marginal makes the residual dominate,  producing the negative conditional values visible in our real-world results.

\item \textbf{Use PID only when Y has interpretable discrete structure.} It is the most precise estimator we tested on clean data (essentially zero unique information for irrelevant text) but requires discretizations, and on continuous real targets the binning choice becomes a free parameter that materially changes the answer.

\item \textbf{Prefer paired contrasts over absolute values.} Comparing conditional MI for a corpus against the same estimator's value for a shuffled or misaligned version of that corpus cancels the estimator's bias, whereas comparing an absolute conditional MI against zero does not.
While conditional MI values alone can help to decide whether to fuse text, this alignment contrast can be used to decide whether the text-time series pairing is trustworthy. We demonstrated this for Time-MMD and FinTexTS, where text carried measurable information that is not tied to its timestamp.
\end{enumerate}
\section{Limitations}
\label{sec:limitations}
\textbf{Synthetic signal design.}
MMTT-Bench uses a sine wave with externally imposed constant segments, which creates an unusually clean separation between informative and uninformative annotation points. 
This controlled design is a deliberate choice to enable oracle ground truth that real-world datasets cannot provide: without knowing the true data generation process, there is no principled way to assess whether an estimator is correct. MMTT-Bench fills this gap precisely because it is synthetic. We validate that our estimator evaluation results hold when noise is added to the sine signal, and on seven real-world datasets which, as shown in Appendix~\ref{sec:appendix-real-world-datasets}, cover the full spectrum of time series characteristics.


\textbf{Representation Pipeline}
As outlined in section~\ref{sec:representation}, the MI ordering holds between text annotations when varying both the number of lookback steps and PCA text dimensions, and the exact values are also relatively stable across eight different text embedding models. 
An embedding model fine-tuned on text and time series pairs may perform better, but would make it impossible to isolate  \(I(X_\text{text}; Y \mid X_\text{ts})\), the unique contribution of the text without time series, and simple text embedding models represent the realistic deployment setting for annotation auditing in practice, where computational cost constrains embedding choice.

\textbf{Neural estimator instability on weak signals.}
MINE and InfoNCE require sufficient signal strength to train a reliable critic. On real-world datasets, where conditional MI is small, MINE collapses to near-zero or negative estimates, whereas InfoNCE inflates, and both show high variance, making them unreliable in low-gain settings. Hyperparameter optimization was performed across different batch sizes and learning rates, with results in Appendix~\ref{sec:appendix-sine-sweeps},\ref{sec:appendix-real-params}, but with only marginal improvements, demonstrating the fundamental sensitivity to signal strength is a structural limitation of critic-based estimation.

\textbf{PID discretizations.}
Discrete $Y$-binning into unique values is semantically motivated for the sine signal in MMTT-Bench, where only four \(Y\) values are present, but no natural partition exists for continuous real-world targets such as those in Time-MMD. This limits PID to signals with interpretable discrete structure in the target variable, or requires arbitrary binning choices that introduce sensitivity to hyperparameters. Appendix~\ref{sec:appendix-timemmd-params} goes into details of different PID parameters tested.
\vspace{-8pt}
\section{Conclusion}
\label{sec:conclusions}
\vspace{-8pt}
We compare six mutual information estimators on MMTT-Bench, the first synthetic benchmark for evaluating information-theoretic metrics in text-time series multimodal forecasting. 
CCA, V-Information and PID are able to identify correct annotations as more informative on both MMTT-Bench and real-world datasets, whereas neural estimators (MINE, InfoNCE) are unreliable in settings with low conditional MI. The pre-training correlation between metric values and downstream MSE across multiple model architectures validates their use as an annotation auditing tools and fusion selection diagnostics.

These findings lead us to present seven recommendations for when and how to utilize MI estimators in multimodal settings. We release MMTT-Bench generation code, estimator implementations, model training pipelines, and the dataset itself to support this research.

\section*{Ethics statement}
This work introduces tools for evaluating the quality of textual annotations in multimodal time series forecasting before any model is trained. The primary societal benefit is improved reliability and transparency in domains where text-augmented forecasting is consequential: public health, energy demand forecasting, financial risk assessment and environmental monitoring. In adversarial settings, an actor with knowledge of which annotations carry high MI with a target variable could craft annotations that appear informative to MI estimators while encoding misleading signals - the incorrect annotation category studied in this paper demonstrates that anti-correlated text can score above genuinely irrelevant text, a property that could be exploited deliberately. We consider these risks low in the near term, as the methods require access to the time series and future values to compute MI, limiting their applicability to retrospective rather than prospective manipulation. Broader misuse of multimodal forecasting systems in high-stakes automated decision-making remains a concern independent of this work, and we encourage practitioners to treat MI metrics as one component of a broader human-in-the-loop quality assurance process.

All data is synthetically generated and contains no personal, sensitive or proprietary information. The annotation templates and word banks are constructed to described abstract signal properties only, and carry no cultural, political, or demographic content. The Time-MMD and FinTexTS datasets are publically available and used in accordance with their terms.
\section*{Reproducibility statement}
Regarding our synthetic dataset MMTT-Bench, we release the full generation code alongside the dataset on both GitHub\footnote{\code} and HuggingFace\footnote{https://huggingface.co/datasets/WhenDoesTextInform/MMTT-Bench} so that the construction process is fully auditable and the dataset can be regenerated, modified, or extended by the community. Appendix~\ref{sec:appendix-dataset-generation} goes into details about how the dataset was generated, including the templates for all text annotations. 

Our mutual information estimation implementation is also released as part of our code repository, with full mathematical details available in Appendix~\ref{sec:appendix-metrics} and estimator parameters recorded per dataset in Appendix~\ref{sec:appendix-sine-sweeps} and~\ref{sec:appendix-real-params}. Our training code for all fourteen model architectures is also reproducible from our code base, with training parameters and compute requirements detailed in Appendix~\ref{sec:appendix-exp-details}. 

To generate all results in this paper, a single script is provided in our code repository. Further scripts enable regeneration of all tables and plots. As MMTT-Bench is available within the repository itself, these experiments can be reproduced by anyone in a single command. For MMTT-Bench signal perturbations and real-world datasets, we provide references to the exact source data where appropriate and data preprocessing scripts, as well as another a 'additional experiments' script to run all experiments and generate results. 

\section*{Acknowledgments}
We thank Keane Ong for his help and advice throughout the process of writing this paper.

\bibliography{iclr2027_conference}
\bibliographystyle{iclr2027_conference}

\appendix
\section{Dataset Generation Details}
\label{sec:appendix-dataset-generation}
To generate the dataset, or inspect the full code, we make the code available at:\code

First a sine wave of 512 periods is generated, with annotation points every \(\pi/4\). A random mask of periods with length between \([\pi/4, 7\pi/4]\) and total length equal to 30\% of the full signal is applied, with all masked values set to -0.5, creating ``transition'' regions. The final signal is then divided into train (62.5\%), validation (18/75\%) and test (18/75\%) and saved as separate files.

Annotation points are placed at quarter-cycle intervals \(\Delta t = \pi/2\) to align with sine-wave phases, which produces a sparse dataset, realistic with real-world scenarios where human annotation is expensive. 
Each point in assigned a phase based on it's position in the sine period, \texttt{peak}, \texttt{descending\_zero}, \texttt{trough} and \texttt{ascending\_zero}, and they are used as the ground truth so text annotations know the exact phase of the next point being predicted.
These points are assigned phases \texttt{drop\_to\_constant} for annotation points starting immediately before the signal transitions and continuing up until the annotation point immediately before the signal resumes oscillation, which is assigned the \texttt{recover\_from\_constant} phase. 

Correct, incorrect and irrelevant text annotations are generated at each point. For non-transition regions where the signal follows a standard sine wave,the same templates are used for both correct and incorrect text, with all time words being associated to the future and direction and magnitude choices based on the gradient at that annotation point, where steep is defined as gradients exceeding \(\pm 0.7\). These `non-transition` templates can be found below.

\begin{lstlisting}[language=Python, caption={MMTT-Bench, Non-transition templates}, showstringspaces=false]
TEMPLATES = [
    # adverb-medial:
    "The {signal} {time_w} {verb} {adverb}.",
    # imperative-like:
    "Expect the {signal} to {verb} {adverb}.",
    # fronted time clause:
    "{time_clause}, the {signal} {time_w} {verb} {adverb}.",
    # nominal subject:
    "A {adj} {noun} {time_w} characterise the signal.",
    # passive perception:
    "The {signal} {time_w} be observed to be {gerund} {adverb}.",
    # existential:
    "There {time_w} be a {adj} {noun} in the signal.",
    # nominal predicate:
    "The {signal}'s behaviour {time_w} take a {adj} {noun}.",
    # participial absolute:
    "{gerund} {adverb}, the {signal} {time_w} continue past this point.",
    # fronted clause + nominal:
    "{time_clause}, a {adj} {noun} is evident.",
    # wh-nominal subject:
    "What characterises the {signal} {time_w} be its {adj} {noun}.",
]

SIGNAL_WORDS = [
    "sine wave", "sinusodial signal", "sinusoid",
    "simple harmonic motion signal", "sine curve",
    "sinusoid waveform", "sinus wave", "flux signal",
    "pure tone", "harmonic wave"
]

TIME_WORDS = {
    "future":  ["will", "is about to", "is expected to",
                "will soon", "is going to"],
}

TIME_CLAUSES = {
    "future":  ["Going forward", "From this point on",
                "In the near term", "Looking ahead"],
}

DIRECTION = {
    "increasing": {
        "verb":    ["rise", "increase", "climb", "ascend", "grow"],
        "noun":    ["rise", "increase", "ascent", "climb", "upward movement"],
        "gerund":  ["rising", "increasing", "climbing", "ascending", "growing"],
    },
    "decreasing": {
        "verb":    ["fall", "decrease", "drop", "descend", "decline"],
        "noun":    ["fall", "decrease", "descent", "drop", "downward movement"],
        "gerund":  ["falling", "decreasing", "dropping", "descending", "declining"]
    },
}

MAGNITUDE = {
    "steep":   {
        "adverb": ["steeply", "sharply", "rapidly",
                   "significantly", "substantially"],
        "adj":    ["steep", "sharp", "rapid", "significant", "substantial"],
    },
    "shallow": {
        "adverb": ["gently", "gradually", "slowly", "modestly", "slightly"],
        "adj":    ["gentle", "gradual", "slow", "modest", "slight"],
    },
    "neutral": {
        "adverb": ["", "noticeably", "measurably"],
        "adj":    ["", "noticeable", "measurable"],
    },
}

IRRELEVANT_TEMPLATES = [
    "The signal {phrase}.",
    "This function {phrase}.",
    "The waveform {phrase}.",
    "Notably, this signal {phrase}.",
    "In terms of its global properties, the signal {phrase}.",
    "What can be observed is that the waveform {phrase}.",
    "A feature worth noting is that the signal {phrase}.",
    "It is the case that this function {phrase}.",
    "The measured signal {phrase}.",
    "{phrase} - this is a property of the signal.",
]

IRRELEVANT_PHRASES = {
    "periodicity": [
        "repeats after a fixed interval",
        "completes a full cycle periodically",
        "returns to its previous value after one period",
        "oscillates with a constant frequency",
        "exhibits periodic behaviour",
    ],
    "boundedness": [
        "remains bounded between its minimum and maximum values",
        "does not exceed its amplitude",
        "is confined within a fixed range",
        "stays within a fixed interval at all times",
        "has a finite amplitude",
    ],
    "smoothness": [
        "is continuously differentiable everywhere",
        "has no discontinuities",
        "varies smoothly at every point",
        "has a well-defined derivative at this location",
        "changes without any abrupt transitions",
    ],
    "domain": [
        "is defined for all real-valued inputs",
        "has a domain spanning all real numbers",
        "is well defined at every point along the axis",
        "takes real values across its entire domain",
        "can be evaluated at any point",
    ],
}
\end{lstlisting}

At transition points, the transition templates below are used for correct annotations. Incorrect annotations are generated for the whole signal at once, but with a 30\% chance of using a transition template instead of a non-transition template, so the same number of correct and incorrect points use this template, but the correct ones are only found are \texttt{drop\_to\_constant} and \texttt{recover\_from\_constant} phases, whereas the incorrect transition templates are randomly distributed throughout the signal.

If incorrect templates only used transition templates at transition points, the incorrect annotations would represent a directly conflicting signal which should provide the exact same amount of information as the correct annotations, making the two categories identical. By distributing transition vocabulary randomly throughout the incorrect annotations, but also maintaining a majority that do directly conflict the signal, the incorrect annotations as a whole provide more information for future prediction than the irrelevant templates, with no directionality at all, but still less that the correct annotations, enabling more fine-grained evaluation of mutual information estimators.
\begin{lstlisting}[language=Python, caption={MMTT-Bench, Transition templates}, showstringspaces=false]

DROP_WORD_BANKS = {
    "verb":   ["cut", "clamp", "collapse", "go", "transition"],
    "noun":   ["cut", "clamp", "collapse", "transition", "descent"],
    "gerund": ["cutting", "clamping", "collapsing", "going", "transitioning"],
}

RECOVERY_WORD_BANKS = {
    "verb":   ["recover", "resume", "return", "emerge", "revive"],
    "noun":   ["recovery", "resumption", "return", "emergence", "revival"],
    "gerund": ["recovering", "resuming", "returning", "emerging", "reviving"],
}

DROP_TEMPLATES = [
    # 0 - simple
    "The signal will no longer follow {signal} and instead soon {verb}"
    "to a negative constant.",
    # 1 - imminent
    "The signal is not a {signal} and is about to {verb} to -0.5.",
    # 2 - expectation
    "Expect the signal to change from a {signal} and {verb} to negative"
    "horizontal line shortly.",
    # 3 - nominal
    "A {noun} to a constant value is imminent due to a change of signal"
    "from a {signal}.",
    # 4 - gerund-led
    "{gerund} to minus 0.5, the signal will shortly become flat instead"
    "of following a {signal}.",
    # 5 - existential
    "A new signal, different to {signal}, will shortly be {noun} to"
    "negative half.",
    # 6 - signal's behaviour
    "The signal's behaviour will change from {signal} and shortly"
    "become a {noun} to minus half.",
    # 7 - temporal clause
    "In the near term, the signal will {verb} to - 0.5 rather than"
    "following a {signal}.",
    # 8 - wh-nominal
    "What will characterise the signal next is no longer a {signal}"
    "but a {noun} to a flatline.",
    # 9 - passive-adjacent
    "The signal is not a {signal} anymore and will shortly be observed"
    "to {verb} to a flat period.",
]

RECOVERY_TEMPLATES = [
    # 0
    "The signal will soon {verb} from -0.5 back to a {signal}.",
    # 1
    "The signal is about to {verb} its oscillation as a {signal}.",
    # 2
    "Expect the signal to {verb} from its flat period shortly.",
    # 3
    "A {noun} from negative half to follow a {signal} is imminent.",
    # 4
    "{gerund} from a negative horizontal line, the {signal} will shortly"
    "oscillate again.",
    # 5
    "There will shortly be a {noun} from minus half to a {signal}.",
    # 6
    "The signal's behaviour will shortly become a {signal}, {noun} from negative.",
    # 7
    "In the near term, the {signal} will {verb} from its flat period.",
    # 8
    "What will characterise the {signal} next is a {noun} from constant value.",
    # 9
    "The {signal} will shortly be observed to {verb} from negative 0.5.",]

\end{lstlisting}

\section{Information Theoretic Metric Details}
\begin{table}[h]
\caption{Comparison of mutual information estimators used in the benchmark. 
\(\dagger\) InfoNCE is a lower bound with ceiling log N nats (4.16 nats at \(N=64)\). \(\S\) KSG conservative bound \(d\leq\sqrt{N/2}\), liberal bound \(d \leq N/5\).\label{tab:estimators}}

\centering
\small
\begin{tabularx}{1.\textwidth}{XXXXXXX}
\toprule\
 & \textbf{KSG}  & \textbf{MINE} & \textbf{InfoNCE} & \textbf{CCA} & \textbf{V-info} & \textbf{PID} \\
\midrule
\textbf{Quantity} & I(X;Y) & I(X;Y)& I(X;Y) & I(X;Y) & \(\Delta R^{2}\) & R,U,S\\
\textbf{Estimate type} & Asymp.\ exact  & Lower bound & Lower bound\(^\dagger\)  & Exact (Gaussian) & Exact (linear) & Exact (discrete) \\
\textbf{Input type} & Continuous & Continuous & Continuous & Continuous & Continuous & Discrete \\
\textbf{Assumption} & None & None & None  & Gaussian & Linear & None\\
\textbf{Dim. limit} & \(d\leq \sqrt{N/2}\)\(^\S\)
 & None & None  & \(N> d_X + d_Y\) & \(d \ll N\) & \(C^{d_X} \cdot C^{d_Y}\)\\
\textbf{Decompos.} & No  & No & No & No & No & Yes\\
\textbf{Rep.-dependent} & No  & No & No & No & Yes & No\\
\textbf{Chain rule} & Approx. & Approx. & Approx. & Exact & Exact & N/A \\
\textbf{Reference} & \citep{ksg2004}  & \citep{mine2018} & \citep{oord2018} & \citep{murphy2023} & \citep{xu2020} & \citep{liang2023quantifying}\\
\bottomrule
\end{tabularx}
\end{table}

We implement and evaluate six estimators, summarized in Table~\ref{tab:estimators}. 
\label{sec:appendix-metrics}
\subsection{K-nearest-neighbor estimator}
\label{sec:ksg}
We estimate mutual information between continuous representations using the k-nearest-neighbor estimator (KSG) of~\citet{ksg2004}. 
For two random variables with \(N\) samples, KSG estimates \(I(X;Y)\) by exploiting the relationship between entropy and nearest-neighbor distances, avoiding explicit density estimation. 
For each sample \(i\), let \(\epsilon_{i}\) denote the Chebyshev distance to the \(k\)-th nearest neighbor in the joint space $(X,Y)$. 
The marginal neighbor counts \(n_{X}^{(i)}\) and \(n_{Y}^{(i)}\) are then obtained by querying the corresponding marginal spaces within the same radius. The estimator is
\begin{equation}
    \hat{I}(X;Y) = \psi(k) + \psi(N) - \left\langle \psi(n_X + 1) \right\rangle - \left\langle \psi(n_Y + 1) \right\rangle,
\end{equation}
where \(\psi\) denotes the digamma function and \( \left\langle . \right\rangle\) denotes the empirical mean\citep{ksg2004}. 
Rather than using bootstrapping to estimate variance like all other metrics, we follow \citep{holmes2019estimation}'s corrected variance estimator based on \(1/N\) scaling of KSG variance, due to the known overestimation of MI in bootstrapped samples\citep{holmes2019estimation}.

The reliability of KSG estimates is characterized by two sample-size bounds, a conservative bound \(d \leq \sqrt{N/2}\) and a liberal bound \(d \leq N/5\), beyond which estimates become unreliable~\citep{gao2017}. 

\subsection{Partial Information Decomposition}
\label{sec:pid}
Mutual information estimates of \(I(X_{\mathrm{text}};Y)\) conflate the unique contributions of text with information redundantly shared with the time series. To disentangle these interactions we apply Partial Information Decomposition \citep{williams2010}, which decomposes the total joint information \(I(X_{\mathrm{ts}},X_{\mathrm{text}};Y)\) into four non-negative atoms following~\citet{liang2023quantifying}: 
\begin{equation}
    I(X_{\mathrm{ts}}, X_{\mathrm{text}}; Y) = R + U_{\mathrm{ts}} + U_{\mathrm{text}} + S,
\end{equation}
where R is \textit{redundancy}, i.e., information both modalities share about $Y$, \(U_{\mathrm{ts}}\) and \(U_{\mathrm{text}}\) are \textit{unique} contributions of each modality, and \(S\) is \textit{synergy}, i.e., information available only from their combination. 
Following~\citet{bertschinger2014}, redundancy is defined as the solution to a convex optimization over the set of joint distributions consistent with the observed marginals, which we solve using the CVXPY implementation of \citet{liang2023quantifying}. 

Since PID requires discrete inputs, we discretize each modality prior to estimation. 
Text embeddings are first reduced by PCA to \(d_{\mathrm{text}}\) dimensions and then clustered into \(C_{\mathrm{text}}\) discrete labels via \(k\)-means. 
Time series patch features are similarly clustered into \(C_{\mathrm{ts}}\) labels. 

\subsection{V-Usable Information}
\label{sec:v-info}
Mutual information measures statistical dependence in the distribution irrespective of whether a specific model can exploit it. \citet{xu2020} propose \emph{predictive \(V\)-information} as a complementary quantity that incorporates the computational constraints of the observer: given a predictive family \(V\), the \(V\)-entropy \(H_{V}(Y|X)\) is the minimum cross-entropy achievable by any function in family \(V\), where \(V\)-information is defined as
\begin{equation}
    I_\mathcal{V}(X \to Y) = H_\mathcal{V}(Y) - H_\mathcal{V}(Y \mid X).
\end{equation}
Without constraints on \(V\), \(V\)-information recovers Shannon mutual information. Under a linear Gaussian predictive family with squared loss, \(V\)-information specializes to the coefficient of determination \citep{xu2020}: 
\begin{equation}
    I_\mathcal{V}(X \to Y) = R^2(X \to Y) = 1 - \frac{\mathrm{Var}(Y - \hat{Y})}{\mathrm{Var}(Y)},
\end{equation}
where \(\hat{Y}\) is the prediction of a Ridge regression fit on \(X\). 
We estimate the conditional \(V\)-information of text beyond the time series as 
\begin{equation}
    I_\mathcal{V}(X_\text{text} \to Y \mid X_\text{ts}) = R^2(X_\text{ts}, X_\text{text}) - R^2(X_\text{ts}),
\end{equation}
which measures the gain in explained variance when text embeddings are added to the time series features. 
Unlike KSG conditional MI, this quantity is reliable at small sample sizes and high embedding dimensionality up to the number of samples, since Ridge regression with regularization \(\alpha\) is well-conditioned even when the number of features approaches the number of samples. 
\(V\)-information is representation-dependent by construction, as it measures information that a linear model can exploit from the given embedding, not the Shannon-theoretic information in the underlying distribution. 

\subsection{Neural Mutual Information Estimators}
\label{sec:neural}
We additionally implement two neural estimators: MINE and InfoNCE. Both estimators compute four quantities per annotation category, namely \(I(X_{\mathrm{text}};Y), I(X_{\mathrm{ts}};Y)\), \(I(X_{\mathrm{ts}},X_{\mathrm{text}};Y)\) and \(X_{\mathrm{text}};Y|X_{\mathrm{ts}}\), using the same subtraction \(\hat{I}_{\mathrm{joint}} - \hat{I}_{\mathrm{ts}}\) for the conditional as KSG, with both terms estimated from the same data resample for consistency. 
Unlike KSG, both joint and marginal problems are estimated in the same representational space by the same critic architecture, so the bias difference that causes chain-rule violations in KSG does not arise.

\subsubsection{MINE}
\label{sec:mine}
The Mutual Information Neural Estimator of \citep{mine2018} rewrites MI as a Kullback-Leibler divergence and exploits its Donsker-Varadhan representation \citep{donsker1983} to obtain a lower bound estimable by gradient descent:
\begin{equation}
    I(X;Y) \geq \sup_{T \in \mathcal{F}} \mathbb{E}_{p(x,y)}\left[T(x,y)\right] - \log \mathbb{E}_{p(x)p(y)}\left[e^{T(x,y)}\right],
\end{equation}
where \(T\) is a statistics network parameterized by a two-layer MLP trained with Adam to maximize the bound. We utilize an open-source MINE implementation\footnote{\url{https://github.com/gtegner/mine-pytorch/tree/master}}:
Negative samples are drawn from the product of marginals by shuffling \(Y\) within the batch, and an exponential moving average stabilization of the denominator is used to reduce gradient variance \citep{mine2018}, training for 500 iterations per MI estimate. 
MINE is not sensitive to the curse of dimensionality in the way KSG is, since the critic learns a scalar score function rather than estimating a density in $d$-dimensional space.

\subsubsection{InfoNCE}
\label{sec:infonce}
The InfoNCE estimator of~\citet{oord2018} frames MI estimation as a \(N\)-way classification problem.
For a batch of \(N\) pairs \((x_{i},y_{i}\) drawn from the joint distribution \(p(x,y)\), the critic \(T\) scores all \(N^2\) combinations \((x_{i},y_{i}\) and the estimator is:
\begin{equation}
    \hat{I}_{\text{NCE}}(X;Y) = \mathbb{E}\left[\frac{1}{N}\sum_{i=1}^{N} \left( T(x_i, y_i) - \log \sum_{j=1}^{N} e^{T(x_i, y_j)} \right)\right] + \log N,
\end{equation}
where the \(\log N\) term converts the biased \(\log N\) normalization to an unbiased lower bound~\citep{oord2018}.
The estimator is trained identically to MINE, a two-layer MLP critic optimized with Adam for 500 iterations, but constructs negatives implicitly from the off-diagonal entries of \(N \times N\) score matrix rather than explicitly shuffling marginals. 
InfoNCE provides a tighter bound than MINE in practice but saturates at log \(N\) nats.

\subsection{Canonical Correlation Analysis Estimator}
\label{sec:cca}
As a model-based complement to the distribution-free estimators above, we implement a mutual information estimator based on Canonical Correlation Analysis (CCA), following \citet{murphy2023} (Ch.28), and included as a reference estimator in \citet{beyondnormal2023}. 
Under the assumption that the joint distribution \(p(x,y)\) is multivariate Gaussian, MI admits the closed-form expression:
\begin{equation}
    I(X; Y) = -\frac{1}{2} \sum_{k} \log(1 - \rho_k^2),
\end{equation}
where \(\rho_1, \ldots, \rho{\min(d_X,d_Y)}\) are the canonical correlations between \(X\) and \(Y\), obtained as the singular value of \(L_{XX}^{-1} C_{XY} L_{YY}^{-\top}\) with \(L_{XX}\) and \(L+{YY}\) the Cholesky factors of the regularized marginal covariance matrices \(C+{XX} + \varepsilon I\) and \(C+{YY} + \varepsilon I\) respectively. 

Unlike KSG, MINE and InfoNCE, CCA requires no hyperparamter tuning, no nearest-neighbor searches and no neural network training, it reduces to a single matrix decomposition and is therefore deterministic given a bootstrap resample. 
\citet{beyondnormal2023} find that CCA achieves the lowest sample complexity of all estimators they evaluate, and remains competitive even when the Gaussianity assumption is mildly violated, though it fails on heavy-tailed and strongly non-linear distributions such as spiral embeddings. 
For our benchmark signals, CCA provides a reliable upper bound on the linear-Gaussian MI against which the distribution-free estimators can be calibrated. The chain rule holds exactly under the Gaussian assumption, so CCA conditional MI estimates do not exhibit the chain-rule inconsistency that causes KSG to produce negative conditional MI.

\section{Embedding Visualizations}
\label{sec:appendix-embed}
\begin{figure}[h]
  \centering
  \includegraphics[width=\textwidth]{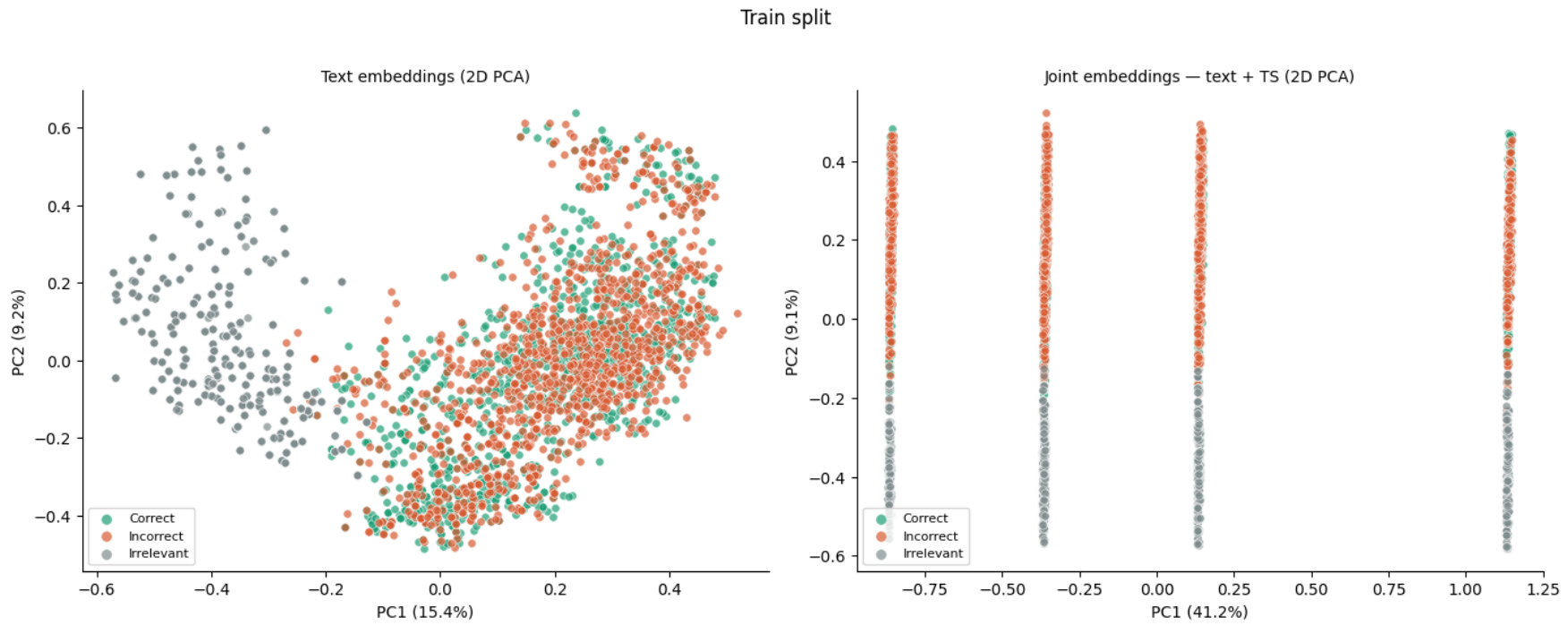}
  \caption{Text embeddings for sine wave training dataset}
  \label{fig:sine_embed}
\end{figure}

Figure~\ref{fig:sine_embed} shows 2D PCA of the text-only embeddings \(X_{\mathrm{text}}\) on the left, and text + time series joint embeddings \(X_{\mathrm{joint}}\) on the right. 
Correct and incorrect text use identical template structures, as detailed in Appendix~\ref{sec:appendix-dataset-generation}, so are undifferentiable unless joined with a time series, at which point they either support or contradict the future prediction.

Indeed, the text embedding (left plot) show a correct (green dots)-incorrect (orange dots) overlap, while the irrelevant annotation cluster (gray dots) is distinctly separated from the correct-incorrect cluster. The \(X_{joint} = [X_{ts};X_{text}]\) embeddings in Figure~\ref{fig:sine_embed} show a dominant structure (PC1, 41.2\% of variance) reflecting the four discrete time series states \(\{-1,0,+1,-0.5\}\), visible as four vertical stripes. 
Text categories are almost entirely superimposed within each stripe, confirming the signal from the correct text is a fine-grained conditional effect relative to the dominant time series structure.

\section{Experiment Details}
\label{sec:appendix-exp-details}
All experiments were conducted on a server equipped with 2xAMD EPYC 7763 64-core processors (128 physical cores, 256 logical CPUs), 2 TiB RAM, and 4xNVIDIA GeForce RTX 4090 GPUs (24 GB VRAM each), running Ubuntu 22.04.5 LTS with CUDA driver 570.133.07.

Experiments were generally lightweight and fast to run, apart from the MINE estimator which on MMTT-Bench training split of \(N=1279\) points takes \(\approx53\pm0.64\)seconds per MI estimate when running only on CPU, where our experiment runs 20 estimates for each result. 
\subsection{Model Training}
We trained two classes of models for both MMTT-Bench and real world datasets: five simple scikit-learn regressors and nine time-series transformer architectures with ten fusion strategies each.
\subsubsection{Scikit-learn regressors}
Five models, each fit on the concatenation of the time-series lookback window and the PCA-reduced text embedding, with the following defaults:
ridge regression (\(\alpha = 1.0\), MLP (hidden-layers 128-64, max 500 iterations), SVR (RBF kernel, C=10, \(\varepsilon = 0.01\)), gradient boosting (200 estimators, max depth 4, learning rate 0.05), random foresst (200 estimators, max depth 8), and k-nearest-neighbors (k=10, distance-weighted). Seeds are fixed at 42 where the model is stochastic. The joint feature vector is a horizontal concatenation of the two modalities, deliberately the simplest possible fusion, so that any gain is attributable to the information in the text rather than to a learned fusion mechanism. 

\subsubsection{Transformer backbones}
We used the Controlled Fusion Adapter (CFA) implementation from~\citep{lee2026rethinking}, which extends time series transformer backbones to accept a text context. We train nine architectures: PatchTST, DLinear, iTransformer, Autoformer, FEDformer, Informer, TiDE, FiLM and Nonstationary Transformer. Shared hyperparameters follow CFA's Table E.1~\citep{lee2026rethinking}. Training uses batch size 32, up to 30 epochs with early stopping at patience 5, learning rate \(1e^{-3}\) for the backbone and \(1e^{-2}\) for the text projection MLP, with one seed (2021). Text is encoded with BERT (6 layers, average pooling, frozen) and projected by an MLP from \(d_{\text{llm}}\) to \(d_{\text{llm}}\)/8 to the prediction length, following CFA's `exp\_long\_term\_forecasting\_text\_integrated`. Real datasets use~\citet{timemmd}'s MM-TSFlib sliding-window loader with a StandardScaler. The synthetic dataset is one sample per annotation point, with no window, to preserve the sample independence MI estiamates require. 

\subsubsection{Fusion Strategies}
Each backbone is trained under ten injection odes, all of them from the CFA codebase, spanning three families:
\begin{table}[h]
\centering
\caption{Text injection modes. Every backbone is trained under all eleven, giving
$9 \times 10 = 90$ text-consuming configurations per dataset; the unimodal baseline is
trained once per backbone and shared across them. All modes are taken from the CFA
implementation of~\citet{lee2026rethinking}}
\label{tab:fusion-modes}
\small
\begin{tabularx}{\linewidth}{lXp{0.55\linewidth}}
\toprule
Mode & Source & What it does \\
\midrule
\multicolumn{3}{@{}l}{\emph{Baseline}} \\
\texttt{unimodal} & CFA & Text is ignored entirely; the no-text baseline every comparison is made against. \\
\addlinespace
\multicolumn{3}{@{}l}{\emph{Naive fusion}} \\
\texttt{first-additive}  & CFA & Projected text embedding is added to the series representation before the backbone encoder. \\
\texttt{middle-additive} & CFA & Added to the intermediate representation inside the backbone. \\
\texttt{last-additive}   & CFA & Added to the backbone output immediately before the prediction head. \\
\texttt{first-concat}    & CFA & Text is concatenated to the series representation at the input, widening the feature dimension. \\
\texttt{middle-concat}   & CFA & Concatenated at the intermediate representation. \\
\texttt{last-concat}     & CFA & Concatenated at the output, before the head. \\
\addlinespace
\multicolumn{3}{@{}l}{\emph{Constrained fusion}} \\
\texttt{film}       & CFA, after~\citet{perez2018film} & Text produces per-channel scale and shift parameters that modulate the series representation, rather than being summed into it. \\
\texttt{gating}     & CFA & A learned gate decides how much text signal passes through, so the model can suppress it. \\
\texttt{orthogonal} & CFA & The text contribution is projected onto the subspace orthogonal to the series representation, so it can only add information the series does not already carry. \\
\texttt{cfa}        & CFA (proposed method by~\cite{lee2026rethinking}) & Constrained fusion adapter: cross-attention from series to text through a bottleneck (reduction factor 8). \\
\bottomrule
\end{tabularx}
\end{table}
Nine backbones times ten text-consuming models give the 90 configurations summarized in every distribution we report below in Appendix~\ref{sec:appendix-sine-final} and~\ref{sec:appendix-real-results}. The unimodal baseline is trained once per backbone and shared across them. 

\section{MMTT-Bench Parameters}
\label{sec:appendix-sine-sweeps}
Table~\ref{tab:sine_params} shows the final parameters used for MMTT-Bench experiments. In the following sub-sections we report parameter sweeps for the subset of hyperparameters whose optimal values were not directly inherited from the dataset configuration or established conventions in the MI estimation literature.
\begin{table}[h]
\centering
\caption{Final hyperparameters for MMTT-Bench experiments}
\label{tab:sine_params}

\begin{subtable}{\linewidth}
\centering
\caption{Overall hyperparameters}
\begin{tabular}{ll}
\toprule
\textbf{Parameter} & \textbf{Value} \\
\midrule
Embedding model       & \texttt{sentence-transformers/all-distilroberta-v1} \\
Time-series tokeniser & Identity \\
Horizon steps         & 1 \\
Lookback steps        & 2 \\
Patch length          & 1 \\
Stride                & 1 \\
PCA dimension         & 16 \\
Bootstrap resamples   & 20 \\
Embedding strategy    & Discrete \\
Shuffle               & True for shuffle experiment, False for everything else \\
\bottomrule
\end{tabular}
\end{subtable}

\vspace{1em}

\begin{subtable}{\linewidth}
\centering
\caption{Per-estimator hyperparameters}
\begin{tabularx}{\linewidth}{llX}
\toprule
\textbf{Estimator} & \textbf{Parameter} & \textbf{Value} \\
\midrule
KSG            & Neighbors $k$                       & 1 \\
\midrule
MINE           & Training iterations                  & 500 \\
               & Hidden dimension                     & 128 \\
               & Batch size $N$                       & 64 \quad ($\log N \approx 4.16$ nat ceiling) \\
               & Learning rate                        & $10^{-4}$ \\
\midrule
InfoNCE        & Training iterations                  & 500 \\
               & Hidden dimension                     & 128 \\
               & Batch size $N$                       & 64 \quad ($\log N \approx 4.16$ nat ceiling) \\
               & Learning rate                        & $10^{-3}$ \\
\midrule
CCA            & Regularization $\varepsilon$         & $10^{-5}$ \\
\midrule
V-information  & Ridge CV folds                       & 5 \\
\midrule
PID            & Time-series clusters $C_\text{ts}$   & 4 \\
               & Text clusters $C_\text{text}$        & 5 \\
               & Target bins $n_Y$                    & 4 (mirrors the four discrete signal values \(Y \in \{1,0,-1,-0.5\}\)) \\
               & Target channel \(y\)  & 0 \\
\bottomrule
\end{tabularx}
\end{subtable}

\end{table}
\subsection{Text Tokenization and Dimensionality reduction}
Tables~\ref{tab:embed_sweep} and~\ref{tab:pca_sweep}
contain the full results that are summarized in Figure~\ref{fig:representation} in the main manuscript. 
Table~\ref{tab:embed_sweep} compares sentence-transformer models; DistilRoBERTa achieves the highest \(X_{text}\) information relative to \(X_{ts}\) and is used as the baseline.
Table~\ref{tab:pca_sweep} shows as text dimensions increase the information provided by \(X_{text}\) across all estimators also increases. KSG is unreliable above \(d\leq \sqrt{N/2} \approx25\), so we choose \(d_{text}=16\) as the highest viable value. 
\begin{table}[t]
\centering
\caption{MI estimator results for different text embedding models.}
\tiny
\label{tab:embed_sweep}
\begin{tabularx}{\textwidth}{Xcccccc}
\toprule
Model & KSG & MINE & InfoNCE & CCA & V-information & PID (U$_{\text{text}}$) \\
\midrule
glove.840B.300d & 0.135 \(\pm\) 0.068 & 0.282 \(\pm\) 0.056 & 0.521 \(\pm\) 0.028 & 0.141 \(\pm\) 0.014 & 0.140 \(\pm\) 0.011 & 0.196 \(\pm\) 0.014 \\
all-MiniLM-L6-v2 & 0.166 \(\pm\) 0.067 & 0.357 \(\pm\) 0.057 & 0.546 \(\pm\) 0.029 & 0.234 \(\pm\) 0.016 & 0.213 \(\pm\) 0.012 & 0.179 \(\pm\) 0.018 \\
all-distilroberta-v & 0.261 \(\pm\) 0.068 & 0.427 \(\pm\) 0.057 & 0.557 \(\pm\) 0.026 & 0.297 \(\pm\) 0.019 & 0.254 \(\pm\) 0.014 & 0.307 \(\pm\) 0.014 \\
bert-base-uncased & 0.199 \(\pm\) 0.072 & 0.326 \(\pm\) 0.051 & 0.541 \(\pm\) 0.027 & 0.149 \(\pm\) 0.016 & 0.146 \(\pm\) 0.013 & 0.344 \(\pm\) 0.017 \\
gpt2 & 0.268 \(\pm\) 0.067 & 0.345 \(\pm\) 0.053 & 0.546 \(\pm\) 0.027 & 0.128 \(\pm\) 0.026 & 0.128 \(\pm\) 0.013 & 0.186 \(\pm\) 0.014 \\
Llama-2-7b-hf & 0.315 \(\pm\) 0.067 & 0.353 \(\pm\) 0.053 & 0.539 \(\pm\) 0.027 & 0.124 \(\pm\) 0.014 & 0.124 \(\pm\) 0.013 & 0.121 \(\pm\) 0.015 \\
opt-125m & 0.344 \(\pm\) 0.063 & 0.351 \(\pm\) 0.049 & 0.537 \(\pm\) 0.028 & 0.149 \(\pm\) 0.016 & 0.146 \(\pm\) 0.013 & 0.257 \(\pm\) 0.017 \\
phi-2 & 0.366 \(\pm\) 0.066 & 0.338 \(\pm\) 0.054 & 0.531 \(\pm\) 0.018 & 0.140 \(\pm\) 0.016 & 0.139 \(\pm\) 0.013 & 0.237 \(\pm\) 0.016 \\
oracle & 1.114 \(\pm\) 0.031 & 0.608 \(\pm\) 0.043 & 0.618 \(\pm\) 0.023 & 4.680 \(\pm\) 0.013 & 0.568 \(\pm\) 0.020 & 0.796 \(\pm\) 0.012 \\
\bottomrule
\end{tabularx}
\end{table}
\begin{table}[t]
\centering
\caption{MI estimator results for different PCA dimensions}
\label{tab:pca_sweep}
\begin{tabular}{lcccc}
\toprule
Estimator & PCA dimension \(d_{\text{text}}\) & Correct & Incorrect & Irrelevant \\
\midrule
KSG  & 4 &     0.57  &       0.119 &        0.183 \\
& 8 &     0.514 &       0.102 &        0.193 \\
& 16 &     0.261 &      -0.011 &        0.118 \\
& 32 &     0.026 &      -0.199 &        0.011 \\
MINE & 4 &     0.318 &       0.033 &        0.047\\
& 8 &     0.4   &       0.101 &        0.015 \\
& 16&     0.413 &       0.138 &        0.025 \\
& 32 &     0.463 &       0.182 &       -0.019  \\
InfoNCE & 4 &     0.429 &       0.125 &        0.105 \\
& 8 &     0.515 &       0.278 &        0.18 \\
& 16 &     0.557 &       0.417 &        0.244 \\
& 32&     0.58  &       0.507 &        0.3  \\
CCA & 4 &     0.115 &       0.003 &        0.002 \\
& 8 &     0.184 &       0.026 &        0.004 \\
& 16&     0.297 &       0.053 &        0.01 \\
& 32 &     0.349 &       0.075 &        0.022 \\
V-information & 4  &     0.117 &       0.003 &        0.002 \\
& 8  &     0.174 &       0.028 &        0.005 \\
& 16 &     0.254 &       0.057 &        0.011 \\
& 32 &     0.285 &       0.078 &        0.022 \\
PID (U$_{\text{text}}$) & 4 &     0.386 &       0     &        0     \\
& 8 &     0.301 &       0     &        0   \\
& 16 &     0.307 &       0.001 &        0 \\
& 32 &     0.318 &       0.001 &        0 \\
\bottomrule
\end{tabular}
\end{table}
\subsection{Time Series Tokenization}
For time series tokenization, we first test \(X_{ts}\) as \(Y\) values of the previous \(n_{lookback}\) steps. Table~\ref{tab:lookback-sweep} shows an input of \(n_{lookback}=2\) ensures the time series provides some predictive information without making prediction trivial in phases that follow a sine wave oscillation. We did experiment using PatchTST time series tokenization, introduced by~\citep{PatchTST}, with results for different patch and stride lengths available in Table~\ref{tab:patch-sweep}, however given the simplicity of the signal we opt to keep \(X_{ts}\) as raw \(Y\) values.
\begin{table}[t]
\centering
\caption{MI estimator results for different numbers of lookback steps. Here there is no time series tokenization, the lookback steps are just passed directly as the time series vector}
\label{tab:lookback-sweep}
\begin{tabular}{lcccc}
\toprule
Estimator & lookback steps \(n_{\text{lookback}}\) & Correct & Incorrect & Irrelevant \\
\midrule
KSG & 1 &     0.907 &       0.601 &        0.555 \\
                   & 2 &     0.261 &      -0.011 &        0.118 \\
                   & 4 &    -0.857 &      -1.061 &       -0.948 \\
                   & 8 &    -0.781 &      -0.894 &       -0.86  \\
MINE                & 1 &     0.586 &       0.314 &        0.187 \\
                    & 2 &     0.43  &       0.146 &        0.05  \\
                    & 4 &     0.396 &       0.077 &        0.014 \\
                    & 8 &     0.413 &       0.093 &        0.028 \\
InfoNCE             & 1 &     0.744 &       0.581 &        0.293 \\
                    & 2 &     0.557 &       0.417 &        0.244 \\
                    & 4 &     0.49  &       0.342 &        0.255 \\
                    & 8 &     0.414 &       0.287 &        0.223 \\
CCA  & 1 &     0.401 &       0.147 &        0.013 \\
                   & 2 &     0.297 &       0.053 &        0.01  \\
                   & 4 &     0.278 &       0.036 &        0.009 \\
                   & 8 &     0.278 &       0.025 &        0.009 \\
V-information & 1 &     0.55  &       0.254 &        0.022 \\
                  & 2 &     0.254 &       0.057 &        0.011 \\
                   & 4 &     0.207 &       0.033 &        0.007 \\
                   & 8 &     0.18  &       0.019 &        0.006 \\
PID (U$_{\text{text}}$) & 1 &     0.222 &       0.003 &        0.002 \\
                   & 2 &     0.307 &       0.001 &        0     \\
                   &4  &     0.321 &       0.002 &        0.001 \\
                   &8  &     0.324 &       0.002 &        0.001 \\
\bottomrule
\end{tabular}
\end{table}
\begin{table}[h]
\centering
\caption{MI estimator results for different numbers of patch lengths, strides and lookback steps. Here PatchTST time series tokenisation is used, where lookback steps must be greater than or equal to patch length}
\label{tab:patch-sweep}
\begin{tabular}{lcccccc}
\toprule
Estimator & patch length & stride & \makecell{lookback steps\\ \(n_{\text{lookback}}\)} & Correct & Incorrect & Irrelevant \\
\midrule
KSG & 1 & 1 & 2 &     0.842 &       0.578 &        0.593 \\
                   & 2 & 1 & 2 &     0.854 &       0.481 &        0.343 \\
                   & 2 & 2 & 2 &     0.854 &       0.481 &        0.343 \\
                   & 4 & 2 &4  &    -0.857 &      -1.061 &       -0.948 \\
                   & 4 & 4 & 4 &    -0.857 &      -1.061 &       -0.948 \\
MINE& 1 & 1 & 2 &     0.404 &       0.156 &        0.038 \\
                   & 2 & 1 & 2 &     0.762 &       0.352 &        0.016 \\
                   & 2 & 2 & 2 &     0.762 &       0.352 &        0.016 \\
                   & 4 & 2 & 4 &     0.397 &       0.068 &        0.012 \\
                   & 4 & 4 & 4 &     0.393 &       0.079 &        0.019 \\
InfoNCE& 1 & 1 & 2 &     0.557 &       0.417 &        0.244 \\
                   & 2 & 1 & 2 &     0.9   &       0.631 &        0.131 \\
                   & 2 & 2 & 2 &     0.9   &       0.631 &        0.131 \\
                   & 4 & 2 & 4 &     0.49  &       0.342 &        0.255 \\
                   & 4 & 4 &4  &     0.49  &       0.342 &        0.255 \\
CCA  & 1 & 1 & 2 &     0.297 &       0.053 &        0.01  \\
                   & 2 & 1 & 2 &     0.226 &       0.037 &        0.014 \\
                   & 2 & 2 & 2 &     0.226 &       0.037 &        0.014 \\
                   & 4 &2  &4  &     0.278 &       0.036 &        0.009 \\
                   & 4 & 4 & 4 &     0.278 &       0.036 &        0.009 \\
V-information & 1 & 1 & 2 &     0.254 &       0.057 &        0.011 \\
                   & 2 & 1 & 2 &     0.306 &       0.059 &        0.02  \\
                   & 2 & 2 & 2 &     0.306 &       0.059 &        0.02  \\
                   & 4 & 2 & 4 &     0.207 &       0.033 &        0.007 \\
                   & 4 & 4 & 4 &     0.207 &       0.033 &        0.007 \\
PID (U$_{\text{text}}$) & 1 & 1 & 2 &     0.307 &       0.001 &        0     \\
                   & 2 & 1 & 2 &     0.297 &       0.002 &        0.002 \\
                   & 2 & 2 & 2 &     0.297 &       0.002 &        0.002 \\
                   & 4 & 2 & 4 &     0.321 &       0.002 &        0.001 \\
                   & 4 & 4 & 4 &     0.321 &       0.002 &        0.001 \\
\bottomrule
\end{tabular}
\end{table}
%
\subsection{PID discretization}
The target \(Y_{\mathrm{future}}\) is binned into four classes, corresponding to the possible signal values at annotation points \(Y \in [1,0,-1,-0.5]\). The joint distribution \((P(X_{\mathrm{ts}},X_{\mathrm{text}},Y)\) is estimated from the resulting discrete triples and passed to the solver.
\newpage
\section{MMTT-Bench Full Results}
\label{sec:appendix-sine-final}
\subsection{MI Estimation}
\begin{table}[h]
\centering
\caption{Conditional mutual information $I(X_{\text{text}}; Y \mid X_{\text{ts}})$ estimates on MMTT-Bench,
in nats, mean $\pm$ standard deviation over 20 bootstrap resamples.}
\label{tab:sine-conditional-mi}
\begin{tabular}{lccc}
\toprule
Estimator & Correct & Incorrect & Irrelevant \\
\midrule
KSG                     & $0.404 \pm 0.066$ & $0.009 \pm 0.064$ & $0.021 \pm 0.067$ \\
MINE                    & $0.426 \pm 0.057$ & $0.125 \pm 0.049$ & $0.030 \pm 0.055$ \\
InfoNCE                 & $0.504 \pm 0.024$ & $0.219 \pm 0.027$ & $0.111 \pm 0.024$ \\
CCA                     & $0.297 \pm 0.019$ & $0.053 \pm 0.009$ & $0.010 \pm 0.003$ \\
Deep CCA                & $2.291 \pm 0.073$ & $0.581 \pm 0.060$ & $0.229 \pm 0.052$ \\
V-information           & $0.254 \pm 0.014$ & $0.057 \pm 0.009$ & $0.011 \pm 0.004$ \\
PID (U$_{\text{text}}$) & $0.307 \pm 0.014$ & $0.001 \pm 0.001$ & $0.000 \pm 0.000$ \\
\bottomrule
\end{tabular}
\end{table}
\clearpage
\subsection{Model Performance}
\begin{longtable}[c]{llrrrr}
\caption{MMTT-Bench results for five simple architectures and nine different time-series Transformer architectures, with each transformer architecture trained with 10 different fusion strategies\label{tab:sine-transformer}}\\
\toprule
\bf{Model} & \bf{Fusion}  & \bf{No text} & \bf{Correct} & \bf{Incorrect} & \bf{Irrelevant} \\
\midrule
ridge & default & 0.2712 & 0.1462 & 0.2607 & 0.2728 \\
\cline{1-6}
mlp & default & 0.2222 & 0.0967 & 0.2828 & 0.2965 \\
\cline{1-6}
svr & default & 0.2503 & 0.1060 & 0.2660 & 0.2836 \\
\cline{1-6}
gbr & default & 0.2161 & 0.0901 & 0.1995 & 0.2478 \\
\cline{1-6}
rf & default & 0.2160 & 0.0886 & 0.1792 & 0.2331 \\
\midrule
\multirow[m]{10}{*}{Autoformer} & cfa & 0.2167 & 0.2106 & 0.2316 & 0.2510 \\
& film & 0.2167 & 0.0875 & 0.2269 & 0.2278 \\
& first-additive & 0.2167 & 0.1727 & 0.2084 & 0.2191 \\
& first-concat & 0.2167 & 0.2095 & 0.2387 & 0.2790 \\
& gating & 0.2167 & 0.1037 & 0.2213 & 0.2387 \\
& last-additive & 0.2167 & 0.1920 & 0.2576 & 0.2456 \\
& last-concat & 0.2167 & 0.1592 & 0.2050 & 0.2127 \\
& middle-additive & 0.2167 & 0.1695 & 0.2224 & 0.2346 \\
& middle-concat & 0.2167 & 0.2348 & 0.2284 & 0.2219 \\
& orthogonal & 0.2167 & 0.1033 & 0.2198 & 0.2569 \\
\cline{1-6}
\multirow[m]{10}{*}{DLinear} & cfa & 0.2120 & 0.2113 & 0.2113 & 0.2113 \\
& film & 0.2120 & 0.1064 & 0.2090 & 0.2118 \\
& first-additive & 0.2120 & 0.1357 & 0.2056 & 0.2134 \\
& first-concat & 0.2120 & 0.1358 & 0.2130 & 0.2133 \\
& gating & 0.2120 & 0.1009 & 0.2031 & 0.2139 \\
& last-additive & 0.2120 & 0.1258 & 0.2027 & 0.2130 \\
& last-concat & 0.2120 & 0.1237 & 0.2083 & 0.2136 \\
& middle-additive & 0.2120 & 0.1027 & 0.2022 & 0.2125 \\
& middle-concat & 0.2120 & 0.1283 & 0.2149 & 0.2168 \\
& orthogonal & 0.2120 & 0.2106 & 0.2135 & 0.2117 \\
\cline{1-6}
\multirow[m]{10}{*}{FEDformer} & cfa & 0.2096 & 0.2210 & 0.2147 & 0.2086 \\
& film & 0.2096 & 0.0560 & 0.2450 & 0.2069 \\
& first-additive & 0.2096 & 0.0674 & 0.2433 & 0.2035 \\
& first-concat & 0.2096 & 0.0642 & 0.2266 & 0.2430 \\
& gating & 0.2096 & 0.1100 & 0.1986 & 0.2268 \\
& last-additive & 0.2096 & 0.1996 & 0.2040 & 0.2039 \\
& last-concat & 0.2096 & 0.1085 & 0.2104 & 0.1902 \\
& middle-additive & 0.2096 & 0.1821 & 0.2228 & 0.2105 \\
& middle-concat & 0.2096 & 0.1045 & 0.2037 & 0.2088 \\
& orthogonal & 0.2096 & 0.1347 & 0.2451 & 0.2278 \\
\cline{1-6}
\multirow[m]{10}{*}{FiLM} & cfa & 0.4097 & 0.4096 & 0.4096 & 0.4096 \\
& film & 0.4097 & 0.4097 & 0.4097 & 0.4097 \\
& first-additive & 0.4097 & 0.4107 & 0.4107 & 0.4107 \\
& first-concat & 0.4097 & 0.4894 & 0.4894 & 0.4894 \\
& gating & 0.4097 & 0.4097 & 0.4097 & 0.4097 \\
& last-additive & 0.4097 & 0.2245 & 0.3280 & 0.4218 \\
& last-concat & 0.4097 & 0.2163 & 0.3277 & 0.3963 \\
& middle-additive & 0.4097 & 0.4097 & 0.4097 & 0.4097 \\
& middle-concat & 0.4097 & 0.4097 & 0.4097 & 0.4097 \\
& orthogonal & 0.4097 & 0.4096 & 0.4096 & 0.4096 \\
\cline{1-6}
\multirow[m]{10}{*}{Informer} & cfa & 0.2130 & 0.2760 & 0.1908 & 0.2359 \\
& film & 0.2130 & 0.0628 & 0.2071 & 0.2307 \\
& first-additive & 0.2130 & 0.0639 & 0.1870 & 0.1927 \\
& first-concat & 0.2130 & 0.2904 & 0.3194 & 0.3665 \\
& gating & 0.2130 & 0.0639 & 0.1883 & 0.1876 \\
& last-additive & 0.2130 & 0.1419 & 0.2048 & 0.2046 \\
& last-concat & 0.2130 & 0.2203 & 0.2419 & 0.2452 \\
& middle-additive & 0.2130 & 0.0942 & 0.2202 & 0.1947 \\
& middle-concat & 0.2130 & 0.1148 & 0.2031 & 0.2038 \\
& orthogonal & 0.2130 & 0.0710 & 0.2231 & 0.2234 \\
\cline{1-6}
\multirow[m]{10}{*}{\shortstack[l]{Nonstationary\\Transformer}} & cfa & 0.2292 & 0.1859 & 0.2071 & 0.2121 \\
& film & 0.2292 & 0.0626 & 0.1970 & 0.1993 \\
& first-additive & 0.2292 & 0.0573 & 0.2064 & 0.2022 \\
& first-concat & 0.2292 & 0.4011 & 0.4096 & 0.1893 \\
& gating & 0.2292 & 0.0405 & 0.1795 & 0.1953 \\
& last-additive & 0.2292 & 0.1526 & 0.2113 & 0.1780 \\
& last-concat & 0.2292 & 0.1392 & 0.1743 & 0.1666 \\
& middle-additive & 0.2292 & 0.0632 & 0.2131 & 0.1699 \\
& middle-concat & 0.2292 & 0.1955 & 0.1990 & 0.1916 \\
& orthogonal & 0.2292 & 0.0566 & 0.1928 & 0.2187 \\
\cline{1-6}
\multirow[m]{10}{*}{PatchTST} & cfa & 0.2785 & 0.2727 & 0.2763 & 0.2740 \\
& film & 0.2785 & 0.2244 & 0.3013 & 0.2805 \\
& first-additive & 0.2785 & 0.2734 & 0.2734 & 0.2734 \\
& first-concat & 0.2785 & 0.2734 & 0.2734 & 0.2734 \\
& gating & 0.2785 & 0.3194 & 0.3368 & 0.4090 \\
& last-additive & 0.2785 & 0.2150 & 0.2700 & 0.2620 \\
& last-concat & 0.2785 & 0.1994 & 0.2429 & 0.2712 \\
& middle-additive & 0.2785 & 0.2789 & 0.2500 & 0.2985 \\
& middle-concat & 0.2785 & 0.3009 & 0.3126 & 0.3604 \\
& orthogonal & 0.2785 & 0.2343 & 0.2622 & 0.3478 \\
\cline{1-6}
\multirow[m]{10}{*}{TiDE} & cfa & 0.2295 & 0.2325 & 0.2325 & 0.2325 \\
& film & 0.2295 & 0.2330 & 0.2330 & 0.2330 \\
& first-additive & 0.2295 & 1977.9946 & 2383.0618 & 1956.9061 \\
& first-concat & 0.2295 & 0.1448 & 0.2246 & 0.2288 \\
& gating & 0.2295 & 0.2318 & 0.2318 & 0.2318 \\
& last-additive & 0.2295 & 0.1075 & 0.2123 & 0.2319 \\
& last-concat & 0.2295 & 0.1340 & 0.2218 & 0.2267 \\
& middle-additive & 0.2295 & 0.2324 & 0.2324 & 0.2324 \\
& middle-concat & 0.2295 & 0.2329 & 0.2329 & 0.2329 \\
& orthogonal & 0.2295 & 0.2305 & 0.2305 & 0.2305 \\
\cline{1-6}
\multirow[m]{10}{*}{iTransformer} & cfa & 0.1840 & 0.2090 & 0.1762 & 0.2088 \\
& film & 0.1840 & 0.0417 & 0.1836 & 0.1940 \\
& first-additive & 0.1840 & 0.0394 & 0.1747 & 0.1810 \\
& first-concat & 0.1840 & 0.0408 & 0.1837 & 0.1926 \\
& gating & 0.1840 & 0.0776 & 0.2043 & 0.1945 \\
& last-additive & 0.1840 & 0.1210 & 0.2062 & 0.1902 \\
& last-concat & 0.1840 & 0.1046 & 0.1905 & 0.1728 \\
& middle-additive & 0.1840 & 0.0445 & 0.2040 & 0.2018 \\
& middle-concat & 0.1840 & 0.0579 & 0.1776 & 0.2001 \\
& orthogonal & 0.1840 & 0.0607 & 0.2321 & 0.2218 \\
\cline{1-6}
\bottomrule
\end{longtable}

\subsection{Signal Perturbation Experiments}
\label{appendix:signal-perturbations}
\subsubsection{Shuffle Results}
\label{sec:appendix-shuffle}
To confirm that the incorrect signal is carried by the text-time series pairing rather than the text alone, we shuffle all annotations across time points and recompute all metrics. Table~\ref{tab:shuffle} shows full results for MINE and InfoNCE,  which both drop their in MI estimates significantly after shuffling, with no change in irrelevant text, confirming that the inverted signal is a consistent property of the (text, \(Y_{\mathrm{future}})\) pairing and not an artifact of embedding distribution.
\begin{table}[h]
\caption{Comparison of conditional MI before and after dataset shuffle
}
\label{tab:shuffle}
\centering
\small
\begin{tabularx}{0.58\textwidth}{XXXXX}
\toprule
\textbf{Category}  & \multicolumn{2}{c}{\textbf{MINE}} & \multicolumn{2}{c}{\textbf{InfoNCE}}  \\
 & Pre-Shuffle & Post-Shuffle & Pre-Shuffle & Post-Shuffle \\
\midrule
Correct & 0.56\(\pm\)0.06 & 0.08\(\pm\)0.05& 0.64\(\pm\)0.02 & 0.16\(\pm\)0.02 \\
Incorrect & 0.17\(\pm\)0.06 & 0.12\(\pm\)0.05 & 0.24\(\pm\)0.02 & 0.17\(\pm\)0.02 \\
Irrelevant & 0.04\(\pm\)0.05 & 0.05\(\pm\)0.04 & 0.12\(\pm\)0.02 & 0.12\(\pm\)0.02 \\
\bottomrule
\end{tabularx}
\end{table}
\vspace{-2ex}
\subsubsection{Signal Perturbations}
Our sine signal in MMTT-Bench is not representative of real data by construction, because the information content of every annotation is known, and it is this property that makes estimator evaluator possible.
That said, MMTT-Bench generation is extensible, and we demonstrate that estimators still rank correct text highest with a noisy sine signal at four different noise levels in Table~\ref{tab:noisy_sine} 
\begin{table}[h]
\caption{Conditional MI $I(X_{\text{text}}; Y \mid X_{\text{ts}})$ (nats, mean \(\pm\) std)
  on the noisy sine benchmark. Observation noise $\sigma$ is added to the signal, annotations are unchanged.
}
\label{tab:noisy_sine}
\centering
\small
\begin{tabular}{ccccc}
\toprule
Estimator & $\sigma$ & Correct & Incorrect & Irrelevant \\
\midrule
\multirow{4}{*}{KSG} & 0.05 & \textbf{0.067 \(\pm\) 0.028} & -0.267 \(\pm\) 0.032 & -0.210 \(\pm\) 0.038 \\
 & 0.1 & \textbf{0.019 \(\pm\) 0.035} & -0.288 \(\pm\) 0.042 & -0.223 \(\pm\) 0.038 \\
 & 0.2 & \textbf{0.087 \(\pm\) 0.035} & -0.156 \(\pm\) 0.037 & -0.144 \(\pm\) 0.042 \\
 & 0.4 & \textbf{0.143 \(\pm\) 0.036} & -0.046 \(\pm\) 0.042 & -0.110 \(\pm\) 0.047 \\
\midrule
\multirow{4}{*}{MINE} & 0.05 & \textbf{0.433 \(\pm\) 0.047} & 0.135 \(\pm\) 0.060 & 0.020 \(\pm\) 0.056 \\
 & 0.1 & \textbf{0.475 \(\pm\) 0.063} & 0.128 \(\pm\) 0.048 & 0.045 \(\pm\) 0.053 \\ 
 & 0.2 & \textbf{0.482 \(\pm\) 0.065} & 0.217 \(\pm\) 0.051 & 0.089 \(\pm\) 0.057 \\
 & 0.4 & \textbf{0.451 \(\pm\) 0.042} & 0.347 \(\pm\) 0.041 & 0.185 \(\pm\) 0.049 \\
\midrule
\multirow{4}{*}{InfoNCE} & 0.05 & \textbf{0.067 \(\pm\) 0.029} & 0.453 \(\pm\) 0.036 & 0.286 \(\pm\) 0.022 \\
 & 0.1 & \textbf{0.771 \(\pm\) 0.036} & 0.501 \(\pm\) 0.038 & 0.312 \(\pm\) 0.035 \\ 
 & 0.2 & \textbf{0.804 \(\pm\) 0.027} & 0.624 \(\pm\) 0.038 & 0.411 \(\pm\) 0.047 \\
 & 0.4 & \textbf{0.860 \(\pm\) 0.049} & 0.803 \(\pm\) 0.035 & 0.515 \(\pm\) 0.046 \\
\midrule
\multirow{4}{*}{CCA} & 0.05 & \textbf{0.309 \(\pm\) 0.019} & 0.038 \(\pm\) 0.007 & 0.013 \(\pm\) 0.003 \\
 & 0.1 & \textbf{0.283 \(\pm\) 0.022} & 0.060 \(\pm\) 0.007 & 0.009 \(\pm\) 0.003 \\
 & 0.2 & \textbf{0.223 \(\pm\) 0.019} & 0.049 \(\pm\) 0.010 & 0.014 \(\pm\) 0.004 \\
 & 0.4 & \textbf{0.170 \(\pm\) 0.014} & 0.037 \(\pm\) 0.006 & 0.010 \(\pm\) 0.004 \\
\midrule
\multirow{4}{*}{V-Information} & 0.05 & \textbf{0.264 \(\pm\) 0.015} & 0.042 \(\pm\) 0.007 & 0.015 \(\pm\) 0.004 \\
 & 0.1 & \textbf{0.253 \(\pm\) 0.015} & 0.066 \(\pm\) 0.008 & 0.009 \(\pm\) 0.004 \\
 & 0.2 & \textbf{0.228 \(\pm\) 0.015} & 0.059 \(\pm\) 0.011 & 0.017 \(\pm\) 0.006 \\
 & 0.4 & \textbf{0.220 \(\pm\) 0.015} & 0.054 \(\pm\) 0.008 & 0.012 \(\pm\) 0.007 \\
\midrule
\multirow{4}{*}{PID (U$_{\text{text}}$)} & 0.05 & \textbf{0.362 \(\pm\) 0.017} & 0.002 \(\pm\) 0.002 & 0.000 \(\pm\) 0.001 \\
 & 0.1 & \textbf{0.298 \(\pm\) 0.012} & 0.009 \(\pm\) 0.003 & 0.002 \(\pm\) 0.001 \\
 & 0.2 & \textbf{0.152 \(\pm\) 0.011} & 0.005 \(\pm\) 0.002 & 0.006 \(\pm\) 0.003 \\
 & 0.4 & \textbf{0.042 \(\pm\) 0.008} & 0.007 \(\pm\) 0.003 & 0.005 \(\pm\) 0.002 \\
\bottomrule
\end{tabular}
\end{table}
\subsubsection{Temporal Jitter}
We also introduce controlled perturbations in the form of temporal jitters, where each annotation is replaced by the correct text of a nearby time point, so coupling decays while the text stays lexically identically to the correct corpus. Table~\ref{tab:jitter} shows under temporal jitter, estimated MI decays monotonically to the irrelevant-text level as displacement grows. However downstream performance is not monotone, it is instead worst (maximized) at an intermediate displacement and then partially recovers. The interpretation is that slightly-misaligned text is worse than useless, it is locally plausible and therefore actively misleading, whereas heavily-displaced text is simply de-correlated.
\begin{table}[h]
\caption{MI estimates under temporal jitter, where each annotation is replaced by the correct text or a nearby timepoint with \(\sigma\) as the standard deviation of the displacement, in annotation steps.
}
\label{tab:jitter}
\centering
\small
\begin{tabularx}{0.9\textwidth}{Xccccccc}
\toprule
Jitter $\sigma$ & KSG & MINE & InfoNCE & CCA & V-Info & PID & mean MSE \\
\midrule
0 (correct) & 0.199 & 0.419 & 0.557 & 0.308 & 0.261 & 0.393 & 0.100\\
0.5  & 0.027 & 0.260 & 0.477 & 0.139 & 0.138 & 0.277 & 0.211 \\
1 & -0.090 & 0.154 & 0.419 & 0.080 & 0.084 & 0.197 & 0.265 \\
2 & -0.111 & 0.113 & 0.373 & 0.049 & 0.053 & 0.083 & 0.285 \\
4 & -0.130 & 0.050 & 0.369 & 0.032 & 0.035 & 0.029 & 0.314 \\
8 & -0.116 & 0.053 & 0.374 & 0.017 & 0.019 & 0.0004 & 0.291 \\
16 & -0.125 & 0.049 & 0.371 & 0.015 & 0.016 & 0.002 & 0.280 \\
\bottomrule
\end{tabularx}
\end{table}
\section{Real World Datasets}
\label{sec:appendix-real-world-datasets}
Time-MMD~\cite{timemmd} text annotations are sourced from Google web search results, filtered and disentangled from predictions by LLMs and summarized for usability. We construct the TimeMMD datasets using sliding windows based on the approach taken by~\citep{lee2026rethinking}, excluding the Economy and Security domains as these have fewer than 300 training samples per category.

Since Time-MMD contains only real text with no ground truth quality labels, we construct two annotation categories. All processed Time-MMD text is labelled \emph{correct}, acknowledging that it is not verified ground truth but represents the best available domain-relevant information. \emph{Irrelevant} text is constructed by randomly sampling processed text from a different domain, following previous work by~\citet{lee2026rethinking}.

We add an additional real world dataset FinTexTS~\cite{lee2026fintexts}, which combines stock prices with news articles paired based on their semantic embeddings. Due to compute constraints we select a subset of ten stocks: AMD, BA, COST, DIS, GOOGL, INTC, NFLX, NVDA, T and TSLA.

 \begin{table}[t]
\caption{MMTT-Bench and real world dataset feature comparison.
Counts are annotated points per text category. Trend \(R^2\) is the coefficient of determination of an ordinary least-squares fit on time; lag-1 ACF is the autocorrelation of the raw series; seasonal ACF is the autocorrelation of the first-differenced series at the natural period (12 monthly, 52 weekly, 365 daily, 252 trading days), differenced so that persistence is not mistaken for seasonality; variance ratio is Var(second half) / Var(first half) and mean shift is the change in mean between halves in units of the series standard deviation, the two together indicating departure from stationarity.}
\label{tab:real-world-stats}
\begin{center}
\begin{tabularx}{\textwidth}{Xlcccccccc}
\multicolumn{1}{c}{\bf Dataset}  &\multicolumn{1}{c}{\bf Frequency} &\multicolumn{1}{c}{\bf Train} &\multicolumn{1}{c}{\bf Val} &\multicolumn{1}{c}{\bf Test} &\multicolumn{1}{c}{\bf Trend} &\multicolumn{1}{c}{\bf Lag-1} &\multicolumn{1}{c}{\bf Seasonal} &\multicolumn{1}{c}{\bf Var.} &\multicolumn{1}{c}{\bf Mean}
\\ 
  & & & & &\multicolumn{1}{c}{\bf\(R^2\)} &\multicolumn{1}{c}{\bf ACF} &\multicolumn{1}{c}{\bf ACF} &\multicolumn{1}{c}{\bf ratio} &\multicolumn{1}{c}{\bf shift}
\\ 
\toprule
MMTT-Bench & synthetic & 1280 & 384 & 384 & 0.00 & 0.05 & 0.70 & 1.10 & 0.05 \\
\midrule
Agriculture & monthly & 347 & 50 & 99 & 0.86 & 0.99 & 0.00 & 0.45 & 1.53 \\
Climate & monthly & 347 & 50 & 99 & 0.00 & 0.23 & 0.20 & 0.75 & 0.11 \\
Energy & weekly & 1035 & 149 & 295 & 0.77 & 1.00 & 0.07 & 2.27 & 1.71 \\
Public Health & weekly & 972 & 140 & 277 & 0.04 & 0.95 & 0.31 & 0.91 & 0.44 \\
Social Good & monthly & 630 & 90 & 180 & 0.09 & 0.95 & 0.79 & 1.06 & 0.84 \\
Traffic & monthly & 371 & 54 & 106 & 0.88 & 0.96 & 0.95 & 0.62 & 1.67 \\
\midrule
FinTexTS (median of 10) & trading day & 784 & 260 & 260 & 0.83 & 1.00 & 0.00 & 3.20 & 1.56 \\
\end{tabularx}
\end{center}
\end{table}

\subsection{Final Parameters}
\label{sec:appendix-real-params}

\begin{table}
\centering
\caption{Per-estimator hyperparameters}
\begin{tabular}{lll}
\toprule
\textbf{Estimator} & \textbf{Parameter} & \textbf{Value} \\
\midrule
KSG            & Neighbors $k$                       & 1 \\
\midrule
MINE           & Training iterations                  & 500 \\
               & Hidden dimension                     & 128 \\
               & Batch size $N$                       & 256 \quad ($\log N \approx 5.55$ nat ceiling) \\
               & Learning rate                        & $10^{-4}$ \\
\midrule
InfoNCE        & Training iterations                  & 500 \\
               & Hidden dimension                     & 128 \\
               & Batch size $N$                       & 256 \quad ($\log N \approx 5.55$ nat ceiling) \\
               & Learning rate                        & $10^{-3}$ \\
\midrule
CCA            & Regularization $\varepsilon$         & $10^{-5}$ \\
\midrule
V-information  & Ridge CV folds                       & 5 \\
\midrule
PID            & Time-series clusters $C_\text{ts}$   & 4 \\
               & Text clusters $C_\text{text}$        & 16 \\
               & Target bins $n_Y$                    & 4 \\
               & Target channel \(y\)  & 11 \\
\bottomrule
\end{tabular}
\end{table}

\subsubsection{Time-MMD Parameters}
\label{sec:appendix-timemmd-params}
\begin{table}[h]
\centering
\caption{Final hyperparameters for Time-MMD experiment}
\label{tab:timemmd_params}
\begin{tabular}{ll}
\toprule
\textbf{Parameter} & \textbf{Value} \\
\midrule
Embedding model       & \texttt{sentence-transformers/all-distilroberta-v1} \\
Time-series tokenizer & Patch mean \\
Horizon steps         & 12 \\
Lookback steps        & 12 \\
Patch length          & 4 \\
Stride                & 2 \\
PCA dimension         & 16 \\
Bootstrap resamples   & 20 \\
Embedding strategy    & equal-width bins \\
\bottomrule
\end{tabular}
\end{table}
\subsubsection{FinTexTS Parameters}
\label{sec:appendix-fintexts-params}
\begin{table}[h]
\centering
\caption{Final hyperparameters for FinTexTS experiment}
\label{tab:fintexts_params}
\begin{tabular}{ll}
\toprule
\textbf{Parameter} & \textbf{Value} \\
\midrule
Embedding model       & \texttt{answerdotai/ModernBERT-large} \\
Time-series tokenizer & Patch mean \\
Horizon steps         & 3 \\
Lookback steps        & 64 \\
Patch length          & 8 \\
Stride                & 8 \\
PCA dimension         & 5 \\
Bootstrap resamples   & 20 \\
Embedding strategy    & equal-width bins \\
\bottomrule
\end{tabular}
\end{table}
\subsection{Full Results}
\label{sec:appendix-real-results}
\begin{table}[h]
\caption{Alignment (difference in \(I(X_{\text{ts}}, X_{\text{text}}; Y)\) between correctly-paired text and same-domain text drawn from the wrong timestamp), and conditional \(\Delta MI = I(X_{\text{text}}; Y | X_{\text{ts}})\) for correctly-paired text, calculated across seven real-world datasets, and reported alongside \(\sigma_{\text{align}} = \sqrt{(\sigma_{\text{correct}}^2 + \sigma_{\text{incorrect}}^2)}\) and align z, where a large positive means the *pairing* carries information; \(z \approx 0\) means the text is informative in aggregate but not tied to the timestamp it is attached to. Conditional \(\sigma\) is the bootstrap standard deviation of that paired difference, and conditional \(z = \Delta MI / \text{conditional} \sigma\) tests it against zero. FinTexTS z-scores are computed per stock and Stouffer-pooled across the ten analysis stocks.}
\label{tab:real_align}
\centering
\small
\begin{tabularx}{1.\textwidth}{XXcccccc}
\toprule
\bf{Dataset} & \bf{Estimator} & \bf{align \(\Delta\)MI} & \bf{align \(\sigma\)}&   \bf{align z} &   \makecell{\bf{conditional}\\\bf{\(\Delta\)MI} } & \makecell{\bf{conditional}\\\bf{\(\sigma\)}} & \makecell{\bf{conditional}\\ \bf{z}} \\
\midrule
\multirow{6}{*}{Agriculture} & KSG &      0.4361 &    0.0767 &    5.6857 &   -0.192  &  0.0743 &   -2.5854 \\
& MINE           &      0.0055 &    0.1119 &    0.0487 &    0.0098&  0.0976 &    0.1 \\
& InfoNCE        &     -0.0555 &    0.0631 &   -0.8803 &    2.0315 &  0.11   &   18.4614 \\
& CCA            &      0.0591 &    0.0857 &    0.6896 &    0.3622 &  0.0491 &    7.3752 \\
& V-info  &      0.0081 &    0.0097 &    0.8334 &    0.0157 &  0.0038 &    4.1327 \\
& PID \((U_{\text{text}} + S)\)  &      0.0216 &    0.0287 &    0.7531 &    0.1323 &  0.02   &    6.616  \\
\midrule
\multirow{6}{*}{Climate} & KSG       &      0.018  &    0.0873 &    0.2056 &   -0.0605 &  0.0577 &   -1.0496 \\
& MINE           &      0.0296 &    0.0692 &    0.4284 &    0.1673 &  0.058  &    2.8864 \\
& InfoNCE        &     -0.0065 &    0.0298 &   -0.2193 &    0.0838 &  0.0136 &    6.1485 \\
& CCA            &      0.0865 &    0.066  &    1.3122 &    0.3317 &  0.0412 &    8.046  \\
& V-info  &      0.0215 &    0.0188 &    1.141  &    0.074  &  0.0122 &    6.0662 \\
& PID \((U_{\text{text}} + S)\)  &     -0.0284 &    0.0394 &   -0.7204 &    0.1812 &  0.0223 &    8.144  \\
\midrule
\multirow{6}{*}{Energy} &  KSG       &      0.749  &    0.0804 &    9.3142 &   -2.323  &  0.0648 &  -35.8348 \\
& MINE           &      0.0009 &    0.0725 &    0.0118 &    0.0134 &  0.0351 &    0.3817 \\
& InfoNCE        &      0.2057 &    0.0868 &    2.3697 &    0.8293 &  0.0827 &   10.0276 \\
& CCA            &      0.0277 &    0.0964 &    0.2869 &    0.2211 &  0.0234 &    9.4382 \\
& V-info  &      0.003  &    0.0058 &    0.5155 &    0.0058 &  0.0016 &    3.7011 \\
& PID \((U_{\text{text}} + S)\)  &      0.0514 &    0.0137 &    3.7486 &    0.1143 &  0.01   &   11.3978 \\
\midrule
\multirow{6}{*}{Public Health}&  KSG       &      0.5618 &    0.0545 &   10.306  &   -0.7536 &  0.0565 &  -13.3419 \\
& MINE           &      0.0145 &    0.1294 &    0.112  &    0.0554 &  0.0973 &    0.5689 \\
& InfoNCE        &      0.1232 &    0.0549 &    2.2443 &    0.9501 &  0.0609 &   15.6087 \\
& CCA            &      0.1191 &    0.1111 &    1.0717 &    0.3308 &  0.0332 &    9.971  \\
& V-info  &      0.089  &    0.0239 &    3.7306 &    0.1077 &  0.0175 &    6.1539 \\
& PID \((U_{\text{text}} + S)\)  &      0.0814 &    0.0176 &    4.6244 &    0.1546 &  0.0135 &   11.4341 \\
\midrule
\multirow{6}{*}{Social Good} & KSG      &     -0.3016 &    0.0771 &   -3.9104 &   -0.8073 &  0.0549 &  -14.6978 \\
& MINE           &     -0.1378 &    0.1204 &   -1.1445 &   -0.0501 &  0.0519 &   -0.9655 \\
& InfoNCE        &     -0.0531 &    0.1036 &   -0.5121 &    0.4981 &  0.097  &    5.1347 \\
& CCA            &     -0.0719 &    0.1562 &   -0.4602 &    0.1053 &  0.0135 &    7.7821 \\
& V-info  &     -0.0046 &    0.0202 &   -0.2254 &    0.01   &  0.0021 &    4.6964 \\
& PID \((U_{\text{text}} + S)\)  &     -0.0132 &    0.0145 &   -0.9098 &    0.0525 &  0.0077 &    6.7995 \\
\midrule
\multirow{6}{*}{Traffic} & KSG      &      0.1823 &    0.0787 &    2.3173 &   -3.0186 &  0.101  &  -29.8843 \\
& MINE           &     -0.0125 &    0.0758 &   -0.1644 &   -0.1196 &  0.1054 &   -1.135  \\
& InfoNCE        &     -0.0234 &    0.0322 &   -0.7268 &    0.2064 &  0.0335 &    6.1656 \\
& CCA            &      0.0928 &    0.5337 &    0.1738 &    0.3378 &  0.0839 &    4.0287 \\
& V-info  &      0.007  &    0.016  &    0.4366 &    0.0116 &  0.0046 &    2.5317 \\
& PID \((U_{\text{text}} + S)\)&      0.0073 &    0.0317 &    0.2294 &    0.1107 &  0.0235 &    4.7117 \\
\midrule 
\multirow{6}{*}{FinTexTS} & KSG       &      0.0575 &    0.0303 &    6.1426 &   -1.9379 &  0.0395 & -159.504  \\
& MINE           &      0.1013 &    0.1188 &    2.8314 &    0.0023 &  0.0916 &    0.081  \\
& InfoNCE        &     -0.0052 &    0.0765 &   -0.2542 &    0.2334 &  0.0489 &   16.3132 \\
& CCA            &      0.0081 &    0.105  &    0.2496 &    0.0268 &  0.009  &   10.2013 \\
& V-info &      0.0004 &    0.0051 &    0.3364 &    0.0009 &  0.0005 &    6.1402 \\
& PID \((U_{\text{text}} + S)\)  &      0.0132 &    0.0163 &    2.6543 &    0.0992 &  0.0118 &   27.1126 \\
\bottomrule
\end{tabularx}
\end{table}
\begin{table}[h]
\caption{Performance results for real world datasets, constructed with original text (correct), text shuffled to a new timestamp (incorrect) and text from a different domain (irrelevant). Model performance is reported as MSE median and interquartile range across 9 different time series transformer architectures (Autoformer, DLinear, FEDformer, FiLM, Informer, Nonstationary Transformer, PatchTST, TiDE, iTransformer), each with 10 fusion strategies (cfa, film, first-additive, first-concat, gating, last-additive, last-concat, middle-additive, middle-concat, orthogonal).
MSE is used as the performance metric to align better with other works using these datasets. We report median reduction in MSE versus no-text (positive = text improves). }
\label{tab:real_performance}
\centering
\small
\begin{tabularx}{0.75\textwidth}{cccc}
\toprule
\bf{Dataset} &  \bf{Correct} & \bf{Incorrect} & \bf{Irrelevant} \\
\midrule
Agriculture & -2.5\% [-12.2, +4.0] & -1.2\% [-10.0, +4.2] & -1.7\% [-9.1, +2.5] \\
Climate & +1.9\% [-1.0, +4.7] & +1.6\% [-1.4, +4.9] & +1.4\% [-1.1, +4.2] \\
Energy & -6.7\% [-27.8, +2.9] & -7.8\% [-25.5, +3.9] & -7.0\% [-34.7, +3.5] \\
PublicHealth & -5.1\% [-11.2, +2.6] & -5.2\% [-13.2, +1.9] & -4.3\% [-12.3, +2.0] \\
SocialGood & -1.3\% [-13.1, +9.5] & -0.7\% [-12.8, +9.3] & -2.6\% [-12.1, +5.7] \\
Traffic & -1.8\% [-14.0, +9.6] & -8.1\% [-21.8, -0.5] & -8.2\% [-26.9, +2.7] \\
FinTexTS & -0.6\% [-25.6, +8.1] & -5.1\% [-22.3, +5.6] & -4.8\% [-24.2, +6.5] \\
\bottomrule
\end{tabularx}
\end{table}

\end{document}